%% file: main.tex
\documentclass[10pt,twocolumn,letterpaper]{article}

\usepackage[pagenumbers]{iccv} 

\setcitestyle{square, comma, numbers,sort&compress, super}
\usepackage{enumitem}
\usepackage{algpseudocode}
\usepackage[font=small,labelfont=bf]{caption}
\usepackage{array}
\usepackage{multirow}
\usepackage{booktabs}
\usepackage{algorithm}
\usepackage{subcaption}
\usepackage[normalem]{ulem}
\usepackage{xparse}
\usepackage{pifont}
\usepackage{bm}
\usepackage{threeparttable}
\usepackage{etoolbox}

\usepackage{listings}
\usepackage{mwe} %
\usepackage{makecell}
\usepackage{color, colortbl}
\usepackage{tabularx}
\usepackage{pifont}%
\usepackage[accsupp]{axessibility}
\usepackage[symbol]{footmisc}
\usepackage{pgfplots}
\usetikzlibrary{spy}
\usepackage[scaled=0.85]{DejaVuSansMono}

\input{preamble}

\definecolor{citecolor}{HTML}{0071bc}
\usepackage[breaklinks=true,bookmarks=false,colorlinks,bookmarks=false, citecolor=citecolor, pagebackref]{hyperref}

\usepackage[capitalize]{cleveref}
\crefname{section}{\S}{\S\S}
\crefname{subsection}{\S}{\S\S}
\crefformat{table}{Table~#2#1#3}
\crefformat{figure}{Figure~#2#1#3}
\crefformat{equation}{Eq~#2#1#3}

\def\paperID{3287} %
\def\confName{ICCV}
\def\confYear{2025}

\title{Scaling Laws for Native Multimodal Models}

\author{ \scalebox{0.85}{Mustafa Shukor$^{2}$ \quad \quad Enrico Fini$^{1}$ \quad \quad
    Victor Guilherme Turrisi da Costa$^{1}$ \quad \quad Matthieu Cord$^{2}$} \\ [0.2cm]
    \scalebox{0.86}{Joshua Susskind$^{1}$ \quad \quad Alaaeldin El-Nouby$^{1}$} \\ [0.2cm]
    \scalebox{0.95}{$^{1}$Apple \quad \quad $^{2}$Sorbonne University} \\
}

\begin{document}
\maketitle

\input{sec/0_abstract}    
\input{sec/1_intro}

\input{sec/1.5_preliminaries}

\input{sec/2_method}

\input{sec/2_related_work}

\input{sec/3_conclusion}

{
    \small
    \bibliographystyle{ieeenat_fullname}
    \bibliography{main}
}
 \input{sec/X_suppl}

\end{document}

%% file: preamble.tex
\newlength\savewidth\newcommand\shline{\noalign{\global\savewidth\arrayrulewidth
  \global\arrayrulewidth 1pt}\hline\noalign{\global\arrayrulewidth\savewidth}}

\newlength\thinwidth

\definecolor{Gray}{gray}{0.92}
\definecolor{DarkGray}{gray}{0.5}
\definecolor{LightLateColor}{rgb}{0.88,1,1}
\definecolor{altRowColor}{gray}{0.92}
\definecolor{highlightRowColor}{rgb}{0.9, 0.9, 1}

\newcommand{\grayrow}{\rowcolor{Gray}}

\definecolor{GrayNumber}{gray}{0.5}
\definecolor{GrayXMark}{gray}{0.7}

\expandafter\def\expandafter\normalsize\expandafter{%
    \normalsize%
    \setlength\abovedisplayskip{2pt}%
    \setlength\belowdisplayskip{5pt}%
    \setlength\abovedisplayshortskip{-5pt}%
    \setlength\belowdisplayshortskip{2pt}%
}

\newcommand{\cpar}{\par\noindent\textbf}

\usepackage{pgfplots} %
\pgfplotsset{compat=1.3} %
\usepgfplotslibrary{external} %
\usepgfplotslibrary{groupplots} %
\usepgfplotslibrary{statistics} %
\usetikzlibrary{spy}

\definecolor{LateGradStart}{HTML}{185D79}
\definecolor{LateGradEnd}{HTML}{7AC5E9}

\definecolor{EarlyGradStart}{HTML}{fd4b2f}
\definecolor{EarlyGradEnd}{HTML}{f19e18}

\definecolor{LateColor}{HTML}{1F78B4}
\definecolor{EarlyColor}{HTML}{D32F2F}

\definecolor{MOEGradStart}{HTML}{0a3431}
\definecolor{MOEGradEnd}{HTML}{43c197}

\definecolor{CustomA}{HTML}{f1c40f}
\definecolor{CustomB}{HTML}{d68910}
\definecolor{CustomC}{HTML}{c75f25} %
\definecolor{CustomD}{HTML}{2980b9} %

\definecolor{DarkOrange}{HTML}{c75f25}
\definecolor{Brown}{HTML}{d68910}
\definecolor{Gold}{HTML}{b89d53}
\definecolor{Blue}{HTML}{2980b9}

\definecolor{Blue}{HTML}{0072B2}
\definecolor{Orange}{HTML}{D55E00}

\definecolor{Purple}{HTML}{6A3D9A}
\definecolor{Gold}{HTML}{E6AB02}

\definecolor{Teal}{HTML}{117A65}
\definecolor{Coral}{HTML}{E67E22}

\definecolor{NavyBlue}{HTML}{253494}
\definecolor{SoftYellow}{HTML}{FDD835}

\definecolor{DarkGreen}{HTML}{1B5E20}

\definecolor{SteelGray}{HTML}{37474F}
\definecolor{AquaBlue}{HTML}{26C6DA}

\definecolor{Burgundy}{HTML}{7B1FA2}
\definecolor{SkyBlue}{HTML}{4FC3F7}

\definecolor{OliveGreen}{HTML}{558B2F}
\definecolor{Tangerine}{HTML}{FF9800}

\definecolor{MidnightBlue}{HTML}{1A237E}
\definecolor{Peach}{HTML}{FFAB91}

\definecolor{CustomA}{HTML}{f1c40f}

\definecolor{CustomA_Light3}{HTML}{f5d15d} %
\definecolor{CustomA_Light2}{HTML}{f9e282} %
\definecolor{CustomA_Light1}{HTML}{fce94f} %

\definecolor{CustomA_Base}{HTML}{f1c40f} %

\definecolor{CustomA_Dark1}{HTML}{e1b10f} %
\definecolor{CustomA_Dark2}{HTML}{d09e0f} %
\definecolor{CustomA_Dark3}{HTML}{b88a0e} %

\definecolor{CustomG_Light3}{HTML}{77b55c} %
\definecolor{CustomG_Light2}{HTML}{7dbf65} %
\definecolor{CustomG_Light1}{HTML}{85d373} %

\definecolor{CustomG_Base}{HTML}{218838} %

\definecolor{CustomG_Dark1}{HTML}{1b6b2e} %
\definecolor{CustomG_Dark2}{HTML}{17562a} %
\definecolor{CustomG_Dark3}{HTML}{124627} %

\definecolor{CustomC_Light3}{HTML}{c97f4d} %
\definecolor{CustomC_Light2}{HTML}{d98e60} %
\definecolor{CustomC_Light1}{HTML}{e89e73} %

\definecolor{CustomC_Base}{HTML}{b75b20} %

\definecolor{CustomC_Dark1}{HTML}{9c4a1d} %
\definecolor{CustomC_Dark2}{HTML}{8b4219} %
\definecolor{CustomC_Dark3}{HTML}{7b3a15} %

\definecolor{CustomD_Light3}{HTML}{46a8d4} %
\definecolor{CustomD_Light2}{HTML}{63b6de} %
\definecolor{CustomD_Light1}{HTML}{7ac5e9} %

\definecolor{CustomD_Base}{HTML}{2980b9} %

\definecolor{CustomD_Dark1}{HTML}{1f6a8a} %
\definecolor{CustomD_Dark2}{HTML}{175d79} %
\definecolor{CustomD_Dark3}{HTML}{134f66} %

\definecolor{CustomP_Light3}{HTML}{b077d4} %
\definecolor{CustomP_Light2}{HTML}{c08ae0} %
\definecolor{CustomP_Light1}{HTML}{d1a0eb} %

\definecolor{CustomP_Base}{HTML}{8e44ad} %

\definecolor{CustomP_Dark1}{HTML}{732d91} %
\definecolor{CustomP_Dark2}{HTML}{5e2377} %
\definecolor{CustomP_Dark3}{HTML}{4b1b5f} %

\pgfplotsset{
    legend early_same_llm style/.style={
        solid,
        mark=*, %
        CustomA_Base, %
        mark options={solid}, %
        line width=1pt, %
        mark size=1.5pt %
    }
}

\pgfplotsset{
    legend early_same_params style/.style={
        solid,
        mark=*, %
        CustomB, %
        mark options={solid}, %
        line width=1pt, %
        mark size=1.5pt %
    }
}

\pgfplotsset{
    legend early style/.style={
        solid,
        mark=*, %
        CustomC_Base, %
        mark options={solid}, %
        line width=1pt, %
        mark size=1.5pt %
    }
}

\pgfplotsset{
    legend early style/.style={
        solid,
        mark=diamond, %
        CustomC_Base, %
        mark options={solid}, %
        line width=1pt, %
        mark size=1.5pt %
    }
}

\pgfplotsset{
    legend early_noconnect style/.style={
        only marks, %
        mark=*, %
        CustomC_Base, %
        mark options={solid}, %
        line width=0pt, %
        mark size=1.5pt, %
    }
}

\pgfplotsset{
    legend early_0_2b style/.style={
        solid,
        mark=*, %
        CustomC_Light1, %
        mark options={solid}, %
        line width=1pt, %
        mark size=1.5pt %
    }
}

\pgfplotsset{
    legend early_0_4b style/.style={
        solid,
        mark=*, %
        CustomC_Light2, %
        mark options={solid}, %
        line width=1pt, %
        mark size=1.5pt %
    }
}

\pgfplotsset{
    legend early_0_9b style/.style={
        solid,
        mark=*, %
        CustomC_Light3, %
        mark options={solid}, %
        line width=1pt, %
        mark size=1.5pt %
    }
}
\pgfplotsset{
    legend early_2_2b style/.style={
        solid,
        mark=*, %
        CustomC_Dark1, %
        mark options={solid}, %
        line width=1pt, %
        mark size=1.5pt %
    }
}

\pgfplotsset{
    legend early_3_3b style/.style={
        solid,
        mark=*, %
        CustomC_Dark2, %
        mark options={solid}, %
        line width=1pt, %
        mark size=1.5pt %
    }
}

\pgfplotsset{
    legend early_8b style/.style={
        solid,
        mark=star, %
        black, %
        mark options={solid}, %
        line width=1pt, %
        mark size=2pt %
    }
}

\pgfplotsset{
    legend early_0_2b style/.style={
        solid,
        mark=pentagon, %
        CustomC_Light1, %
        mark options={solid}, %
        line width=1pt, %
        mark size=1.5pt %
    }
}

\pgfplotsset{
    legend early_0_4b style/.style={
        solid,
        mark=square, %
        CustomC_Light2, %
        mark options={solid}, %
        line width=1pt, %
        mark size=1.5pt %
    }
}

\pgfplotsset{
    legend early_0_9b style/.style={
        solid,
        mark=o, %
        CustomC_Light3, %
        mark options={solid}, %
        line width=1pt, %
        mark size=1.5pt %
    }
}

\pgfplotsset{
    legend early_init style/.style={
        solid,
        mark=triangle, %
        CustomC_Base, %
        mark options={solid}, %
        line width=1pt, %
        mark size=1.5pt %
    }
}

\pgfplotsset{
    legend early_init style_dashed/.style={
        dashed,
        mark=triangle, %
        CustomC_Base, %
        mark options={solid}, %
        line width=1pt, %
        mark size=1.5pt %
    }
}

\pgfplotsset{
    legend early_init_distill style/.style={
        solid,
        mark=triangle, %
        yellow, %
        mark options={solid}, %
        line width=1pt, %
        mark size=1.5pt %
    }
}

\pgfplotsset{
    legend early_init_distill_distdown style/.style={
        solid,
        mark=triangle, %
        red, %
        mark options={solid}, %
        line width=1pt, %
        mark size=1.5pt %
    }
}

\pgfplotsset{
    legend late_init_aimv2 style/.style={
        solid,
        mark=triangle, %
        brown, %
        mark options={solid}, %
        line width=1pt, %
        mark size=1.5pt %
    }
}

\pgfplotsset{
    legend early bars style/.style={
        ybar, %
        bar width=0.4cm, %
        solid,
        mark=none, %
        fill=CustomC_Base, %
        line width=0pt, %
    }
}

\pgfplotsset{
    legend late_noconnect style/.style={
        only marks, %
        mark=*, %
        CustomD_Base, %
        mark options={solid}, %
        line width=0pt, %
        mark size=1.5pt, %
    }
}

\pgfplotsset{
    legend late style/.style={
        solid,
        mark=*, %
        CustomD_Base, %
        mark options={solid}, %
        line width=1pt, %
        mark size=1.5pt %
    }
}
\pgfplotsset{
    legend late_0_2b style/.style={
        solid,
        mark=*, %
        CustomD_Light1, %
        mark options={solid}, %
        line width=1pt, %
        mark size=1.5pt %
    }
}
\pgfplotsset{
    legend late_0_4b style/.style={
        solid,
        mark=*, %
        CustomD_Light2, %
        mark options={solid}, %
        line width=1pt, %
        mark size=1.5pt %
    }
}
\pgfplotsset{
    legend late_0_9b style/.style={
        solid,
        mark=*, %
        CustomD_Light3, %
        mark options={solid}, %
        line width=1pt, %
        mark size=1.5pt %
    }
}
\pgfplotsset{
    legend late_2_2b style/.style={
        solid,
        mark=*, %
        CustomD_Dark1, %
        mark options={solid}, %
        line width=1pt, %
        mark size=1.5pt %
    }
}
\pgfplotsset{
    legend late_3_3b style/.style={
        solid,
        mark=*, %
        CustomD_Dark2, %
        mark options={solid}, %
        line width=1pt, %
        mark size=1.5pt %
    }
}

\pgfplotsset{
    legend late_init style/.style={
        solid,
        mark=triangle, %
        CustomD_Base, %
        mark options={solid}, %
        line width=1pt, %
        mark size=1.5pt %
    }
}

\pgfplotsset{
    legend moe_noconnect style/.style={
        only marks,
        mark=*, %
        CustomG_Base, %
        mark options={solid}, %
        line width=1pt, %
        mark size=1.5pt %
    }
}

\pgfplotsset{
    legend moe style/.style={
        solid,
        mark=*, %
        CustomG_Base, %
        mark options={solid}, %
        line width=1pt, %
        mark size=1.5pt %
    }
}
\pgfplotsset{
    legend moe_0_2b style/.style={
        solid,
        mark=*, %
        CustomG_Light1, %
        mark options={solid}, %
        line width=1pt, %
        mark size=1.5pt %
    }
}
\pgfplotsset{
    legend moe_0_4b style/.style={
        solid,
        mark=*, %
        CustomG_Light2, %
        mark options={solid}, %
        line width=1pt, %
        mark size=1.5pt %
    }
}
\pgfplotsset{
    legend moe_0_9b style/.style={
        solid,
        mark=*, %
        CustomG_Light3, %
        mark options={solid}, %
        line width=1pt, %
        mark size=1.5pt %
    }
}
\pgfplotsset{
    legend moe_2_2b style/.style={
        solid,
        mark=*, %
        CustomG_Dark1, %
        mark options={solid}, %
        line width=1pt, %
        mark size=1.5pt %
    }
}

\pgfplotsset{
    legend moe_shape style/.style={
        solid,
        mark=diamond, %
        CustomG_Base, %
        mark options={solid}, %
        line width=1pt, %
        mark size=1.5pt %
    }
}
\pgfplotsset{
    legend moe_0_2b_shape style/.style={
        solid,
        mark=pentagon, %
        CustomG_Light1, %
        mark options={solid}, %
        line width=1pt, %
        mark size=1.5pt %
    }
}
\pgfplotsset{
    legend moe_0_4b_shape style/.style={
        solid,
        mark=square, %
        CustomG_Light2, %
        mark options={solid}, %
        line width=1pt, %
        mark size=1.5pt %
    }
}
\pgfplotsset{
    legend moe_0_9b_shape style/.style={
        solid,
        mark=o, %
        CustomG_Light3, %
        mark options={solid}, %
        line width=1pt, %
        mark size=1.5pt %
    }
}

\pgfplotsset{
    legend moe_3_3b style/.style={
        solid,
        mark=*, %
        CustomG_Dark2, %
        mark options={solid}, %
        line width=1pt, %
        mark size=1.5pt %
    }
}

\pgfplotsset{
    legend hard style/.style={
        solid,
        mark=*, %
        CustomP_Base, %
        mark options={solid}, %
        line width=1pt, %
        mark size=1.5pt %
    }
}
\pgfplotsset{
    legend hard_0_2b style/.style={
        solid,
        mark=*, %
        CustomP_Light1, %
        mark options={solid}, %
        line width=1pt, %
        mark size=1.5pt %
    }
}
\pgfplotsset{
    legend hard_0_4b style/.style={
        solid,
        mark=*, %
        CustomP_Light2, %
        mark options={solid}, %
        line width=1pt, %
        mark size=1.5pt %
    }
}
\pgfplotsset{
    legend hard_0_9b style/.style={
        solid,
        mark=*, %
        CustomP_Light3, %
        mark options={solid}, %
        line width=1pt, %
        mark size=1.5pt %
    }
}
\pgfplotsset{
    legend hard_2_2b style/.style={
        solid,
        mark=*, %
        CustomP_Dark1, %
        mark options={solid}, %
        line width=1pt, %
        mark size=1.5pt %
    }
}
\pgfplotsset{
    legend hard_3_3b style/.style={
        solid,
        mark=*, %
        CustomP_Dark2, %
        mark options={solid}, %
        line width=1pt, %
        mark size=1.5pt %
    }
}

\pgfplotsset{
    legend late_vision style/.style={
        dotted,
        mark=*, %
        CustomD_Base, %
        mark options={solid}, %
        line width=1pt, %
        mark size=1.5pt %
    }
}

%% file: sec/0_abstract.tex
\begin{abstract}

Building general-purpose models that can effectively perceive the world through
multimodal signals has been a long-standing goal. Current approaches involve
integrating separately pre-trained components, such as connecting vision
encoders to LLMs and continuing multimodal training. While such
approaches exhibit remarkable sample efficiency, it remains an open question
whether such late-fusion architectures are inherently superior. In this work, we
revisit the architectural design of native multimodal models (NMMs)-those
trained from the ground up on all modalities—and conduct an extensive scaling
laws study, spanning 457 trained models with different architectures and training mixtures. Our investigation reveals no
inherent advantage to late-fusion architectures over early-fusion ones, which do not
rely on image encoders or tokenizers. On the contrary, early-fusion
exhibits stronger performance at lower parameter counts, is more
efficient to train, and is easier to deploy. Motivated by the strong performance of the
early-fusion architectures, we show that incorporating Mixture of Experts (MoEs)
allows models to learn modality-specific weights, significantly
benefiting performance.

\end{abstract}

%% file: sec/1_intro.tex
\section{Introduction}
\label{sec:intro}

Multimodality provides a rich signal for perceiving and understanding the world. Advances in vision  
\cite{radford2021learning,oquab2023dinov2,zhai2023sigmoidsiglip,fini2024multimodalaimv2}  
and language models \cite{achiam2023gpt4,team2023gemini,dubey2024llama3}  
have enabled the development of powerful multimodal models that understand language, images, and audio.  
A common approach involves grafting separately pre-trained unimodal models, such as connecting a vision encoder to the input
layer of an
LLM~\cite{laurenccon2024mattersidefics2,shukor2023epalm,alayrac2022flamingo,
xue2024xgenblip3,beyer2024paligemma,wang2024qwen2,liu2024improvedllava,shukor2025smolvla}.

Although this seems like a convenient approach, it remains an open question
whether such late-fusion strategies are inherently optimal for
understanding multimodal signals.  Moreover, with abundant multimodal data available, initializing from unimodal pre-training is potentially detrimental, as it may introduce biases that prevent the model from fully leveraging
cross-modality co-dependancies. An additional challenge is scaling such systems;  each component (e.g., vision encoder, LLM) has its own set of hyperparameters, pre-training data
mixtues, and scaling properties with respect to the amount of data and compute applied. A more flexible architecture might allow the model to dynamically allocate its capacity across modalities, simplifying scaling efforts.

\input{figs/teaser}

In this work, we focus on the scaling properties of native multimodal models trained from the
ground up on multimodal data. We first investigate whether the commonly adopted late-fusion
architectures hold an intrinsic advantage by comparing them to early-fusion
models, which process raw multimodal inputs without relying on dedicated vision encoders.  
We conduct scaling experiments on early and late fusion architectures, deriving scaling laws to predict their performance and compute-optimal configurations. Our findings indicate that late fusion offers no inherent advantage when trained from scratch. Instead, early-fusion models are more efficient and are easier to scale. Furthermore, we observe that native multimodal models follow scaling laws similar to those of LLMs~\cite{hoffmann2022training}, albeit with slight variations in scaling coefficients across modalities and datasets. Our results suggest that model parameters and training tokens should be scaled roughly equally for optimal performance. Moreover, we find that different multimodal training mixtures exhibit similar overall trends, indicating that our findings are likely to generalize to a broader range of settings.

While our findings favor early fusion, multimodal data is inherently
heterogeneous, suggesting that some degree of parameter specialization may still
offer benefits. To investigate this, we explore leveraging Mixture of Experts
(MoEs)~\cite{shazeer2017outrageously}, a technique that enables the model to
dynamically allocate specialized parameters across modalities in a symmetric and
parallel manner, in contrast to late-fusion models, which are asymmetric and
process data sequentially. 
Training native multimodal models with MoEs results in significantly improved
performance and therefore, faster convergence. Our scaling laws for MoEs suggest that scaling number of
training tokens is more important than the number of active parameters. This
unbalanced scaling is different from what is observed for dense models, due to
the higher number of total parameters for sparse models. In addition, Our analysis reveals that experts tend to
specialize in different modalities, with this specialization being particularly
prominent in the early and last layers.

\subsection{Summary of our findings}
Our findings can be summarized as follows: 
\cpar{Native Early and Late fusion perform on par:} Early fusion models trained from scratch perform on par with their late-fusion counterparts, with a slight advantage
to early-fusion models for low compute budgets (\cref{fig:early_vs_late_scaleflops}).
Furthermore, our scaling laws study indicates that the compute-optimal models for
early and late fusion perform similarly as the compute budget increases~(\cref{fig:teaser} Top).
\cpar{NMMs scale similarly to LLMs:} The scaling laws of native multimodal models follow similar laws as
text-only LLMs with slightly varying  scaling exponents depending on the target 
data type and training mixture (\cref{tab:early_vs_late_coeffs}).
\cpar{Late-fusion requires more parameters:} Compute-optimal late-fusion models
require a higher parameters-to-data ratio when compared to early-fusion (\cref{fig:teaser} bottom).
\cpar{Sparsity significantly benefits early-fusion NMMs:} Sparse NMMs exhibit
significant improvements compared to their dense counterparts at the same
inference cost~(\cref{fig:dense_vs_moe_scaledata}). Furthermore, they implicitly learn modality-specific weights
when trained with sparsity~(\cref{fig:tokens_specialization}). In addition,
compute-optimal models rely more on scaling the number of training tokens than
the number of active parameters as the compute-budget grows (\cref{fig:teaser} Bottom).
\cpar{Modality-agnostic routing beats Modality-aware routing for Sparse NMMs:} Training sparse mixture of experts with modality-agnostic routing consistently
outperforms models with modality-aware routing
(\cref{fig:hard_vs_moe_scaledata}).

\vspace{-5pt}

%% file: figs/teaser.tex
\begin{figure}[t!]
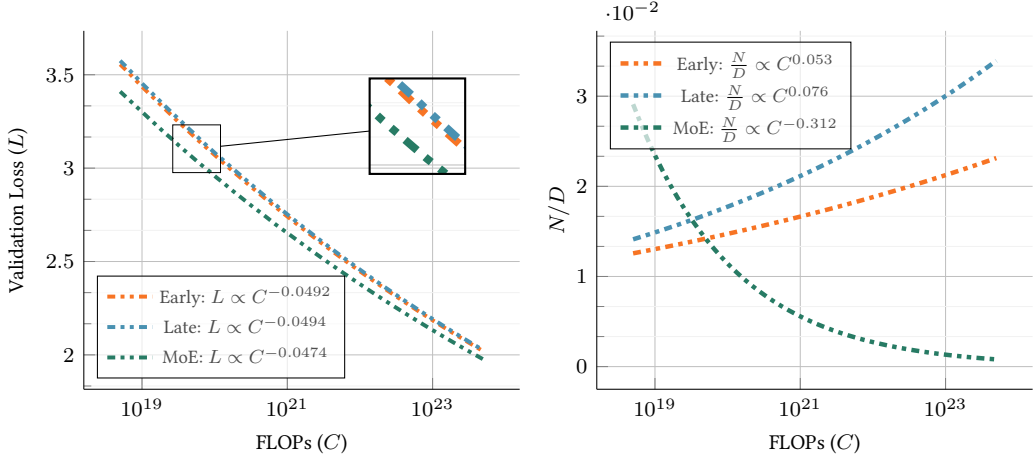

    \centering
    \captionsetup{type=figure}
    \begin{subfigure}[t]{1\linewidth}
        \input{graphs/early_late/loss_vs_flops}
    \end{subfigure}
    \begin{subfigure}[t]{1\linewidth}
        \input{graphs/early_late/d_n_ratio_vs_flops}
    \end{subfigure}
    \caption{\textbf{Scaling properties of Native Multimodal Models.} Based on
    the scaling laws study in \Cref{sec:scaling_laws_early}, we observe: (1)
    early and late fusion models provide similar validation loss $L$ when trained
    with the same compute budget $C$ (FLOPs); (2) This performance is
    achieved via a different trade-off between parameters $N$ and number of
    training tokens $D$, where early-fusion models requires fewer parameters.
    (3) Sparse early-fusion models achieve lower loss and require more
    training tokens for a given FLOP budget.
    }
    \label{fig:teaser}
\end{figure}

%% file: sec/1.5_preliminaries.tex
\section{Preliminaries}

\subsection{Definitions}

\cpar{Native Multimodal Models (NMMs):}
Models that are trained from scratch on all modalities simultaneously without
relying on pre-trained LLMs or vision encoders. Our focus is on the
representative image and text modalities, where the model processes both text
and images as input and generates text as output.

\cpar{Early fusion:} Enabling multimodal interaction from the beginning, using
almost no modality-specific parameters (\eg, except a linear layer to patchify
images). Using a single transformer model, this approach processes raw
multimodal input—tokenized text and continuous image patches—with no
image discretization. In this paper, we refer to the main transformer as
the decoder.

\cpar{Late fusion:} Delaying the multimodal interaction to deeper layers,
typically after separate unimodal components has processed that process each modality
independently (e.g., a vision encoder connected to a decoder).

\cpar{Modality-agnostic routing:} In sparse mixture-of-experts, modality-agnostic routing
refers to relying on a learned router module that is trained jointly with the
model.

\cpar{Modality-aware routing:} Routing based on pre-defined rules such as
routing based on the modality type (\eg, vision-tokens, token-tokens).

\input{tables/notations}

\subsection{Scaling Laws}
We aim to understand the scaling properties of NMMs and how different
architectural choices influence trade-offs. To this end, we analyze our models
within the scaling laws framework proposed by~\citet{kaplan2020scaling,
hoffmann2022training}.  
We compute FLOPs based on the total number of parameters, using the
approximation \(C = 6ND\), as adopted in prior
work~\cite{hoffmann2022training,abnar2025parameters}. However, we modify this
estimation to suit our setup: for late-fusion models, FLOPs is computed as
\(6(N_vD_v + ND)\).  
We consider a setup where, given a compute
budget \(C\), our goal is to predict the model’s final performance, as well as
determine the optimal number of parameters or number of training tokens.
Consistent with prior studies on LLM
scaling~\cite{hoffmann2022training}, we assume a power-law
relationship between the final model loss and both model size (\(N\)) and
training tokens (\(D\)):

\begin{equation}
\label{eq:scaling_laws}
    L = E + \frac{A}{N^{\alpha}} + \frac{B}{D^{\beta}}.
\end{equation}

\noindent Here, \(E\) represents the lowest achievable loss on the dataset,
while \(\frac{A}{N^{\alpha}}\) captures the effect of increasing the number of
parameters, where a larger model leads to lower loss, with the rate of
improvement governed by \(\alpha\). Similarly, \(\frac{B}{D^{\beta}}\) accounts
for the benefits of a higher number of tokens, with \(\beta\) determining the
rate of improvement. Additionally, we assume a linear relationship between
compute budget (FLOPs) and both \(N\) and \(D\) (\(C \propto ND\)). This further
leads to power-law relationships detailed in \cref{tab:power_laws}.

\input{figs/scaling_laws_early_vs_late}

\subsection{Experimental setup}
Our models are based on the autoregressive transformer
architecture~\cite{vaswani2017attention} with SwiGLU FFNs~\cite{shazeer2020glu}
and QK-Norm~\cite{dehghani2023scaling} following~\citet{li2024datacomp}. In
early-fusion models, image patches are linearly projected to match the text
token dimension, while late-fusion follows the CLIP
architecture~\cite{radford2021learning}. We adopt causal attention for text
tokens and bidirectional attention for image tokens, we found this to work
better. Training is conducted on a mixture of public and private multimodal
datasets, including DCLM \cite{li2024datacomp}, Obelics
\cite{laurenccon2024obelics}, DFN \cite{fang2023data}, COYO
\cite{kakaobrain2022coyo700m}, and a private collection of High-Quality
Image-Text Pairs (HQITP). Images are resized to 224×224 resolution with a
14×14 patch size. We use a context length of 1k for the multimodal sequences.
For training efficiency, we train our models with \texttt{bfloat16},  Fully
Sharded Data Parallel (FSDP) \cite{zhao2023pytorch}, activation checkpointing, and gradient accumulation.
We also use sequence packing for the image captioning dataset to reduce the
amount of padded tokens. Similar to previous
works~\cite{hoffmann2022training,aghajanyan2023scalingmm,abnar2025parameters},
we evaluate performance on held-out subsets of interleaved (Obelics),
Image-caption (HQITP), and text-only data (DCLM). Further implementation details
are provided in~\cref{app:implementation_details}.

%% file: tables/notations.tex
\begin{table}[t!]
    \centering
    \setlength{\tabcolsep}{8pt}
    \renewcommand{\arraystretch}{1.0}
    \resizebox{1\linewidth}{!}{
    \begin{tabular}{c p{0.999\linewidth}}
         Expression & Definition  \\
         \shline
         \textbf{$N$}     & \small{{Number of parameters in the multimodal decoder. For MoEs this refers to the active parameters only.}} \\
         \grayrow
         \textbf{$D$}     & \small{{Total number of multimodal tokens.}} \\
         \textbf{$N_{v}$} & \small{{Number of parameters in the vision-specific encoder. Only exists in late-fusion architectures.}} \\
         \grayrow
         \textbf{$D_{v}$} &  \small{{Number of vision-only tokens.}} \\
         \textbf{$C$}     & \small{{Total number of FLOPs, estimated as $C=6ND$ for early-fusion and $C=6(N_vD_v+ND)$ for late-fusion.}} \\
         \grayrow
         \textbf{$L$}     & \small{{Validation loss measured as the average over interleaved image-text, image-caption, and text-only data mixtures.}} \\
    \end{tabular}}
    \caption{Definitions of the expressions used throughout the paper.}
    \label{tab:my_label}
\end{table}

%% file: figs/scaling_laws_early_vs_late.tex
\begin{figure}[t!]
    \centering
    \captionsetup{type=figure}
    \begin{subfigure}[t]{0.48\linewidth}
        \centering
        \includegraphics[width=1.02\linewidth]{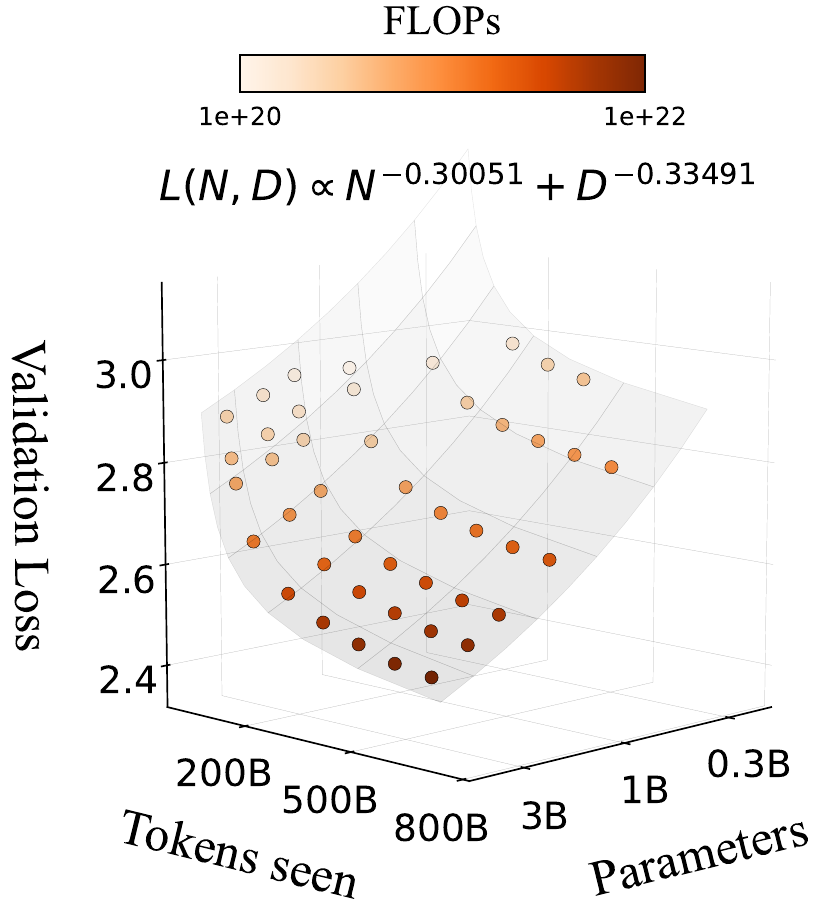}
    \end{subfigure}
    \hfil
    \begin{subfigure}[t]{0.48\linewidth}
        \centering
        \includegraphics[width=1.02\linewidth]{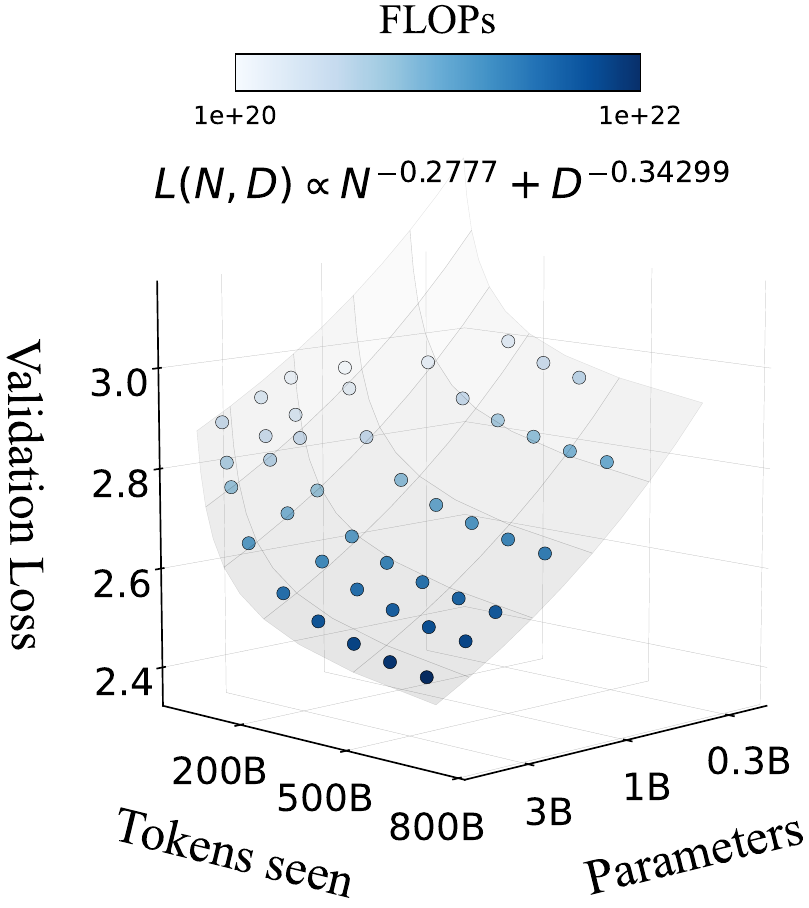}
    \end{subfigure}
    \vspace{5pt}
    \setlength{\fboxsep}{0.5pt}
    \setlength{\fboxrule}{0pt}
    \caption{\textbf{Scaling laws for \fbox{\colorbox{CustomC_Light1!20}{\strut
    early-fusion}} and \fbox{\colorbox{CustomD_Light1!20}{late-fusion\strut}}
    native multimodal models.} Each point represents a model (300M to 3B
    parameters) trained on varying number of tokens (250M to 400B). We
    report the average cross-entropy loss on the validation sets of
    interleaved (Obelics), Image-caption (HQITP), and text-only data
    (DCLM).}
    \label{fig:early_vs_late_scaleflops_3d}
\end{figure}

%% file: sec/2_method.tex
\section{Scaling  native multimodal models}

In this section, we present a scaling laws study of native multimodal models,
examining various architectural choices~\cref{sec:scaling_laws_early}, exploring
different data mixtures~\cref{sec:scaling_data_mix}, analyzing the practical
trade-offs between late and early fusion
NMMs, and comparing the performance of native
pre-training and continual pre-training of NMMs~\cref{sec:native_vs_continual}.  

\cpar{Setup.} We train models ranging from 0.3B to 4B active parameters,
scaling the width while keeping the depth constant. For smaller training token
budgets, we reduce the warm-up phase to 1K steps while maintaining 5K steps for
larger budgets.  
Following~\citet{hagele2024scaling}, models are trained with a constant learning
rate, followed by a cool-down phase using an inverse square root scheduler. The
cool-down phase spans 20\% of the total steps spent at the constant learning
rate.  To estimate the scaling coefficients in \Cref{eq:scaling_laws}, we apply the
L-BFGS algorithm~\cite{lbfgs} and Huber loss~\cite{Huber1992} (with $\delta =
10^{-3}$), performing a grid search over initialization ranges.

\input{tables/scaling_laws_coeffs}

\input{figs/latr_vs_early_equal_flops}

\subsection{Scaling laws of NMMs}
\label{sec:scaling_laws_early}

\cpar{Scaling laws for early-fusion and late-fusion models.}  
\Cref{fig:early_vs_late_scaleflops_3d}~(left) presents the final loss averaged
across interleaved, image-caption, and text datasets for early-fusion NMMs. The
lowest-loss frontier follows a power law as a function of FLOPs. Fitting the
power law yields the expression $L \propto C^{-0.049}$, indicating the rate of
improvement with increasing compute. When analyzing the scaling laws per data
type (\eg, image-caption, interleaved, text), we observe that the exponent
varies (\cref{tab:early_vs_late_coeffs}). For instance, the model achieves a
higher rate of improvement for image-caption data $(L \propto C^{-0.061})$ when
compared to interleaved documents $(L \propto C^{-0.046}$).  

To model the loss as a function of the number of training tokens $D$ and model
parameters $N$, we fit the parametric function in \cref{eq:scaling_laws}, obtaining
scaling exponents $\alpha = 0.301$ and $\beta = 0.335$. These describe the rates
of improvement when scaling the number of model parameters and training tokens, respectively.
Assuming a linear relationship between compute, $N$, and $D$ (\ie, $C \propto ND$),
we derive the law relating model parameters to the compute budget (see
\cref{app:scaling_laws} for details). Specifically, for a given compute budget
$C$, we compute the corresponding model size $N$ at logarithmically spaced $D$
values and determine $N_{opt}$, the parameter count that minimizes loss.
Repeating this across different FLOPs values produces a dataset of $(C,
N_{opt})$, to which we fit a power law predicting the compute-optimal model size
as a function of compute: $N^* \propto C^{0.526}.$

Similarly, we fit power laws to estimate the compute-optimal training dataset
size as a function of compute and model size:  
\[
D_{opt} \propto C^{0.473}, \quad D_{opt} \propto N^{0.899}.
\]  
These relationships allow practitioners to determine the optimal model and
dataset size given a fixed compute budget. When analyzing by data type, we find
that interleaved data benefits more from larger models ($a=0.532$) compared to
image-caption data ($a=0.520$), whereas the opposite trend holds for training
tokens.

We conduct a similar study on late-fusion models
in~\cref{fig:early_vs_late_scaleflops_3d}~(right) and observe comparable scaling
behaviors. In particular, the loss scaling exponent ($c = -0.0494$) is nearly
identical to that of early fusion ($c = -0.0492$).  
This trend is evident in \cref{fig:early_vs_late_scaleflops}, where early fusion
outperforms late fusion at smaller model scales, while both architectures
converge to similar performance at larger model sizes. We also observe similar
trends when varying late-fusion configurations, such as using a smaller vision
encoder with a larger text decoder~\cref{app:late_vs_early}.

\input{figs/late_vs_early_efficiency}

\input{figs/data_mixtures_scaling}

\cpar{Scaling laws of NMMs \textit{vs} LLMs.}
Upon comparing the scaling law coefficients of our NMMs to those reported for
text-only LLMs (\eg, GPT-3, Chinchilla), we find them to be within similar
ranges. In particular, for predicting the loss as a function of compute,
GPT-3~\cite{brown2020language} follows $L \propto C^{-0.048}$, while our models
follow $L \propto C^{-0.049}$, suggesting that the performance of NMMs adheres
to similar scaling laws as LLMs.  
Similarly, our estimates of the $\alpha$ and $\beta$ parameters in
\cref{eq:scaling_laws} ($\alpha=0.301$, $\beta=0.335$) closely match those
reported by~\citet{hoffmann2022training} ($\alpha=0.339$, $\beta=0.285$).
Likewise, our computed values of $a=0.526$ and $b=0.473$ align closely with
$a=0.46$ and $b=0.54$ from~\cite{hoffmann2022training}, reinforcing the idea
that, for native multimodal models, the number of training tokens and model
parameters should be scaled proportionally.  
However, since the gap between $a$ and $b$ is smaller than in LLMs, this
principle holds even more strongly for NMMs. Additionally, as $a=0.526$ is
greater than $b=0.473$ in our case, the optimal model size for NMMs is larger
than that of LLMs, while the optimal number of training tokens is lower, given
a fixed compute budget.

\cpar{Compute-optimal trade-offs for early \textit{vs.} late fusion NMMs.}  
While late- and early-fusion models reduce loss at similar rates with increasing
FLOPs, we observe distinct trade-offs in their compute-optimal models.
Specifically, $N_{opt}$ is larger for late-fusion models, whereas $D_{opt}$ is
larger for early-fusion models. This indicates that, given a fixed compute
budget, late-fusion models require a higher number of parameters, while early-fusion
models benefit more from a higher number of training tokens.  
This trend is also reflected in the lower $\frac{N_{opt}}{D_{opt}} \propto
C^{0.053}$ for early fusion compared to $\frac{N_{opt}}{D_{opt}} \propto
C^{0.076}$ for late fusion. As shown in \cref{fig:teaser}~(bottom), when scaling FLOPs,
the number of parameters of early fusion models becomes significantly lower, which is crucial
for reducing inference costs and, consequently, lowering serving costs after
deployment.

\input{tables/scaling_laws_coeffs_datamixture}

\cpar{Early-fusion is more efficient to train.}
We compare the training efficiency of late- and early-fusion architectures. As shown in \cref{fig:early_vs_late_efficiency}, early-fusion models consume less memory and train faster under the same compute budget. This advantage becomes even more pronounced as compute increases, highlighting the superior training efficiency of early fusion while maintaining comparable performance to late fusion at scale. Notably, for the same FLOPs, late-fusion models have a higher parameter count and higher effective depth (\ie, additional vision encoder layers alongside decoder layers) compared to early-fusion models.

\input{figs/training_mixtures}

\subsection{Scaling laws for different data mixtures}
\label{sec:scaling_data_mix}
We investigate how variations in the training mixture affect the scaling laws of
native multimodal models. To this end, we study four different mixtures that
reflect common community
practices~\cite{laurenccon2024obelics,mckinzie2025mm1,zhang2024mm1_5,lin2024vila},
with Image Caption-Interleaved-Text ratios of \colorbox{blue!10}{45-45-10} (our default setup),
\colorbox{red!10}{30-30-40}, \colorbox{green!10}{40-20-40}, and \colorbox{orange!10}{20-40-40}.  
For each mixture, we conduct a separate scaling study by training 76 different
models, following our setup in \cref{sec:scaling_laws_early}. Overall,
\cref{fig:early_scaleflops_data_mixtures} shows that different mixtures follow
similar scaling trends; however, the scaling coefficients vary depending on the
mixture (\cref{tab:scaling_laws_coeffs_data_mixtures}). Interestingly,
increasing the proportion of image-caption data (mixtures 1 and 2) leads to
lower $a$ and higher $b$, whereas increasing the ratio of interleaved and text
data (mixtures 3 and 4) have the opposite effect.  
Notably, image-caption data contains more image tokens than text
tokens; therefore, increasing its proportion results in more
image tokens, while increasing interleaved and text data increases text token
counts. This suggests that, when image tokens are prevalent, training for longer decreases the loss faster than increasing the model size. 
We also found that for a fixed model size, increasing text-only and interleaved data ratio is in favor of
early-fusion \cref{fig:early_vs_late_datatype_interleaved_text_main}.

\subsection{Native multimodal pre-training \textbf{\vs} continual training of
LLMs}
\label{sec:native_vs_continual}
In this section, we compare training natively from scratch to continual training
after initializing from a pre-trained LLM. We initialize the model from DCLM-1B~\cite{fang2023data} that is trained on more than 2T tokens.
\cref{fig:early_vs_early_init_scaledata} shows that native multimodal models can
close the gap with initialized models when trained for longer.
Specifically, on image captioning data, the model requires fewer than 100B
multimodal tokens to reach comparable performance. However, on interleaved and
text data, the model may need longer training—up to 1T tokens.
Considering the cost of pre-training, these results suggest that training
natively could be a more efficient approach for achieving the same performance on multimodal benchmarks.

\input{figs/early_vs_early_init_scaledata}

\section{Towards multimodal specialization}
Previously, we demonstrated that early-fusion models achieve performance on par with late-fusion models under a fixed compute budget. However, multimodal data is inherently heterogeneous, and training a unified model to fit such diverse distributions may be suboptimal.  
Here, we argue for multimodal specialization within a unified architecture. Ideally, the model should implicitly adapt to each modality, for instance, by learning modality-specific weights or specialized experts. Mixture of Experts is a strong candidate for this approach, having demonstrated effectiveness in LLMs.  
In this section, we highlight the advantages of sparse early-fusion models over their dense counterparts.

\cpar{Setup.} Our sparse models are based on the dropless-MoE implementation of~\citet{gale2023megablocks}, which eliminates token dropping during training caused by expert capacity constraints. We employ a top-$k$ expert-choice routing mechanism, where each token selects its top-$k$ experts among the $E$ available experts. Specifically, we set $k=1$ and $E=8$, as we find this configuration to work effectively.  
Additionally, we incorporate an auxiliary load-balancing loss~\cite{shazeer2017outrageously} with a weight of 0.01 to ensure a balanced expert utilization. Following~\citet{abnar2025parameters}, we compute training FLOPs as $6ND$, where $N$ represents the number of active parameters.

\subsection{Sparse vs dense NMMs when scaling FLOPs}  
We compare sparse MoE models to their dense counterparts by training models with different numbers of active parameters and varying amounts of training tokens. \cref{fig:dense_vs_moe_scaledata} shows that, under the same inference cost (or number of active parameters), MoEs significantly outperform dense models.  
Interestingly, this performance gap is more pronounced for smaller model sizes. This suggests that MoEs enable models to handle heterogeneous data more effectively and specialize in different modalities. However, as dense models become sufficiently large, the gap between the two architectures gradually closes.

\input{figs/scaling_laws_moes}

\subsection{Scaling laws for sparse early-fusion models}
We train different models (ranging from 300M to 3.4B active parameters) on varying amounts of tokens (ranging from 250M to 600B) and report the final loss in \cref{fig:early_scaleflops_moe_avg}. We fit a power law to the convex hull of the lowest loss as a function of compute (FLOPs). Interestingly, the exponent ($-0.048$) is close to that of dense NMMs ($-0.049$), indicating that both architectures scale similarly. However, the multiplicative constant is smaller for MoEs ($27.086$) compared to dense models ($29.574$), revealing lower loss. Additionally, MoEs require longer training to reach saturation compared to dense models (\cref{app:scaling_laws} for more details). We also predict the
coefficients of \cref{eq:scaling_laws} by considering $N$ as the number of
active parameters. \Cref{tab:early_vs_late_coeffs} shows significantly higher
$\alpha$ compared to dense models. Interestingly, $b$ is significantly higher
than $a$, revealing that the training tokens should be scaled at a higher rate
than the number of parameters when training sparse NMMs. We also experiment with a
scaling law that takes into account the sparsity~\citep{abnar2025parameters} and
reached similar conclusions \Cref{app:scaling_laws_moes}.

\subsection{Modality-aware \vs Modality-agnostic routing}

Another alternative to MoEs is modality-aware routing, where multimodal tokens are assigned to experts based on their modalities, similar
to previous works~\cite{bao2021vlmo,wang2022image}. We train models with
distinct image and text experts in the form of FFNs, where image tokens are
processed only by the image FFN and text tokens only by the text FFN. Compared to modality-aware routing, MoEs exhibit significantly better performance on both image-caption and interleaved data as presented in~\cref{fig:hard_vs_moe_scaledata}.

\input{figs/dense_vs_moe_scaledata}

\subsection{Emergence of expert specialization and sharing}
\label{sec:specialization}
We investigate multimodal specialization in MoE architectures. In~\cref{fig:tokens_assignment}, we visualize the normalized number of text and image tokens assigned to each expert across layers.  To quantify this specialization, we compute a specialization score, defined as the average, across all experts within a layer, of $1-H(p)$, where $H$ is the binary entropy of each expert's text/image token distribution. We plot this specialization score in~\cref{fig:tokens_specialization}.  Higher specialization scores indicate a tendency for experts to focus on either text or image tokens, while lower scores indicate a shared behavior.  These visualizations provide clear evidence of modality-specific experts, particularly in the early layers. Furthermore, the specialization score decreases as the number of layers increases, before rising again in the last layers. This suggests that early and final layers exhibit higher modality specialization compared to mid-layers. This behavior is intuitive, as middle layers are expected to hold higher-level features that may generalize across modalities, and consistent with findings in \citep{shukor2024implicit} that shows increasing alignment between modalities across layers. The emergence of both expert specialization and cross-modality sharing in our modality-agnostic MoE, suggests it may be a preferable approach compared to modality-aware sparsity. All data displayed here is from an early-fusion MoE model with 1B active parameters trained for 300B tokens.

\input{tables/sft_results}

\section{Evaluation on downstream tasks with SFT} 
Following previous work on scaling laws, we primarily rely on validation losses. However, we generally find that this evaluation correlates well
with performance on downstream tasks. To validate this, we conduct a multimodal
instruction tuning stage (SFT) on the LLaVA mixture \cite{liu2024improvedllava} and report
accuracy and CIDEr scores across several VQA and captioning tasks.
\cref{tab:sft} confirms the ranking of different model configurations.
Specifically, early fusion outperforms late fusion, and MoEs outperform dense
models. However, since the models are relatively small (1.5B scale), trained
from scratch, and fine-tuned on a small dataset, the overall scores
are lower than the current state of the art.  Further implementation
details can be found in~\Cref{app:implementation_details}.

\input{figs/soft_vs_hard_routing}

%% file: tables/scaling_laws_coeffs.tex
\begin{table}[t!]
    \begin{minipage}[b]{1\linewidth}
        \centering
        \setlength{\tabcolsep}{10pt}
        \renewcommand{\arraystretch}{1.2}
        \resizebox{1\linewidth}{!}{
        \begin{tabular}{c c c c c}
             \shline
             \grayrow $L=E+\frac{A}{N^{\alpha}}+\frac{B}{D^{\beta}}$ & $N \propto C^a$ & $D \propto C^b$ & $L \propto C^c$ & $D \propto N^d$ \\
        \end{tabular}}
      \label{tab:power_laws}
    \end{minipage}
    \begin{minipage}[b]{1\linewidth}
        \centering
        \setlength{\tabcolsep}{3pt}
        \renewcommand{\arraystretch}{1.5}
        \resizebox{1\linewidth}{!}{
        \begin{tabular}{lccccccccc}
            Model & Data & E & $\alpha$ & $\beta$ & a & b  & c & d \\ %
            \shline
            GPT3 \cite{brown2020language} & Text &  -- & -- & -- & -- & -- & -0.048 & \\
            Chinchilla \cite{hoffmann2022training} & Text &  1.693 & 0.339 & 0.285 & 0.46 & 0.54 & -- & \\
            \midrule
            \multirow{4}{*}{NMM (early-fusion)} & Text &  2.222 & 0.3084 & 0.3375 & 0.5246 & 0.4774     & -0.0420 & 0.9085 \\
            & Image-Caption & 1.569 & 0.3111 & 0.3386 & 0.5203 & 0.4785 & -0.0610 & 0.9187 \\
            & Interleaved & 1.966 & 0.2971 & 0.338 & 0.5315 & 0.4680     & -0.0459 & 0.8791 \\
            & AVG & 1.904 & 0.301 & 0.335 & 0.5262 & 0.473 & -0.0492 & 0.8987 \\
            \midrule
            NMM (late-fusion) & AVG &  1.891 & 0.2903 & 0.3383 & 0.6358 & 0.4619 & -0.0494 & 0.6732 \\
            \midrule
            Sparse NMM (early-fusion) & AVG & 2.158  & 0.710 & 0.372 & 0.361  & 0.656 & -0.047 & 1.797 \\
        \end{tabular}%
        } \caption{\textbf{Scaling laws for native multimodal models}. We report the
        scaling laws results for early and late fusion models. We fit the scaling laws for different target data types as well as their average loss (AVG).
        }
        \label{tab:early_vs_late_coeffs}
    \end{minipage}
    
\end{table}

%% file: figs/latr_vs_early_equal_flops.tex
\begin{figure*}[t!]
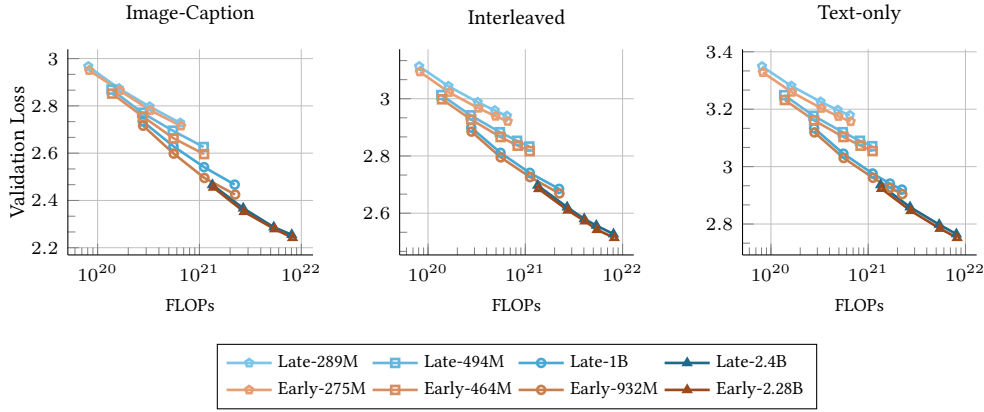

    \centering
    \captionsetup{type=figure}
    \begin{subfigure}[t]{0.33\linewidth}
        \input{graphs/early_late/early_vs_late_scaleflops_getty}
    \end{subfigure}
    \begin{subfigure}[t]{0.33\linewidth}
        \input{graphs/early_late/early_vs_late_scaleflops_obelics}
    \end{subfigure}
    \begin{subfigure}[t]{0.33\linewidth}
    \input{graphs/early_late/early_vs_late_scaleflops_dclm}
    \end{subfigure}
    \vspace{-15pt}
    \begin{center}
        \ref{sharedlegend}
    \end{center}
    \caption{\textbf{Early vs late fusion: scaling training FLOPs.} We compare
    early and late fusion models when scaling both the number of model parameters and the number
    of training tokens. Overall, early fusion shows a slight advantage, especially at smaller model sizes, and the gap decreases when scaling the number of parameters $N$.}
    \label{fig:early_vs_late_scaleflops}
    \vspace{-3mm}
\end{figure*}

%% file: figs/late_vs_early_efficiency.tex
\begin{figure}[t!]
    \centering
    \captionsetup{type=figure}
    \begin{subfigure}[t]{0.49\linewidth}
        \input{graphs/early_late/early_vs_late_scalemodel_mem}
    \end{subfigure}
    \begin{subfigure}[t]{0.49\linewidth}
        \input{graphs/early_late/early_vs_late_scalemodel_time}
    \end{subfigure}
    \vspace{-3pt}
    \caption{\textbf{Early vs late: pretraining efficiency.} Early-fusion is faster to train and consumes less memory. Models are trained on 16 H100
    GPUs for 160k steps (300B tokens).}
    \label{fig:early_vs_late_efficiency}
\end{figure}

%% file: figs/data_mixtures_scaling.tex
\begin{figure*}[h!]
    \centering
    \captionsetup{type=figure}
    \begin{subfigure}[t]{0.24\linewidth}
        \input{graphs/early/early_scaleflops_avg}
    \end{subfigure}
    \begin{subfigure}[t]{0.24\linewidth}
        \input{graphs/early_data_mixtures/early_scaleflops_40_20_40_avg}
    \end{subfigure}
    \begin{subfigure}[t]{0.24\linewidth}
        \input{graphs/early_data_mixtures/early_scaleflops_30_30_40_avg}
    \end{subfigure}
    \begin{subfigure}[t]{0.24\linewidth}
        \input{graphs/early_data_mixtures/early_scaleflops_20_40_40_avg}
    \end{subfigure}
    \begin{tikzpicture}
        \node[anchor=north] (legend) at (0\linewidth, 0) {
            \begin{axis}[
                        hide axis, %
                        xmin=0, xmax=0.5, ymin=0, ymax=1, %
                        legend columns=6, %
                        legend style={
                            at={(-0.12, 0.05)}, %
                            anchor=north, %
                            /tikz/every even column/.append style={column sep=0.2cm}, %
                            scale=0.5,
                            cells={align=left}, font=\scriptsize,
                            anchor=center,
                        },
                    ]
                \addlegendimage{legend early_0_2b style}
                \addlegendentry{0.275B}
                \addlegendimage{legend early_0_4b style}
                \addlegendentry{0.464B}
                \addlegendimage{legend early_0_9b style}
                \addlegendentry{0.932B}
                \addlegendimage{legend early style}
                \addlegendentry{1.627B}
                \addlegendimage{legend early_2_2b style}
                \addlegendentry{2.280B}
                \addlegendimage{legend early_3_3b style}
                \addlegendentry{3.354B}
            \end{axis}
        };
    \end{tikzpicture}
    \caption{\textbf{Scaling laws with different training mixtures.}
    Early-fusion models follow similar scaling trends when changing the pretraining mixtures. However, increasing the image captions leads to a higher scaling exponent norm (see~\cref{tab:scaling_laws_coeffs_data_mixtures}).}
    \label{fig:early_scaleflops_data_mixtures}
    \vspace{-3mm}
\end{figure*}

%% file: tables/scaling_laws_coeffs_datamixture.tex
\begin{table}[t!]
    \centering
    \setlength{\tabcolsep}{3pt}
    \renewcommand{\arraystretch}{1.1}
    \resizebox{\linewidth}{!}{
    \begin{tabular}{lcccccccccc}
        & C-I-T (\%) & I/T ratio &  E & $\alpha$ & $\beta$ & a & b  & d & c \\ %
        \shline
        1 & \colorbox{blue!10}{45-45-10}  & 1.19 & 1.906  & 0.301  & 0.335  & 0.527  & 0.474   & 0.901  & -0.0492 \\
        2 & \colorbox{red!10}{40-20-40}  & 0.65 &  1.965  & 0.328  & 0.348  & 0.518  & 0.486   & 0.937  & -0.0486 \\
        3 & \colorbox{green!10}{30-30-40}  & 0.59 & 1.847  & 0.253  & 0.338  & 0.572  & 0.428   & 0.748  & -0.0463 \\
        4 & \colorbox{orange!10}{20-40-40}  & 0.49 &  1.836  & 0.259  & 0.354  & 0.582  & 0.423   & 0.726  & -0.0488 \\
    \end{tabular}%
    } \caption{\textbf{Scaling laws for different training mixtures}. Early-fusion models. C-I-T refer to image-caption, interleaved and text}
    \label{tab:scaling_laws_coeffs_data_mixtures}
\end{table}

%% file: figs/training_mixtures.tex
\begin{figure}[t!]
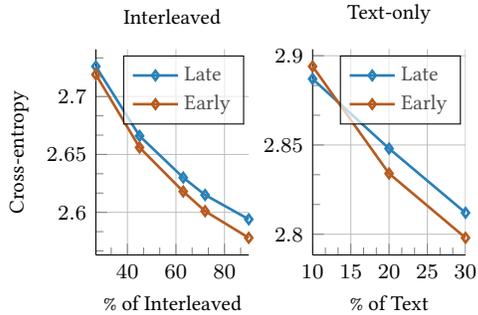

    \centering
    \begin{subfigure}[t]{0.49\linewidth}
        \input{graphs/early_late/early_late_datatype_sameflops_obelics}
    \end{subfigure}
    \begin{subfigure}[t]{0.49\linewidth}
        \input{graphs/early_late/early_vs_late_textratio_dclm}
    \end{subfigure}
    \vspace{-5pt}
    \captionof{figure}{\textbf{Early vs late fusion: changing the training mixture.} We
    vary the training mixtures and plot the final training loss. Early fusion
    models attain a favorable performance when increasing the proportion of interleaved documents
    and text-only data.}
    \label{fig:early_vs_late_datatype_interleaved_text_main}
\end{figure}

%% file: figs/early_vs_early_init_scaledata.tex
\begin{figure}[t!]
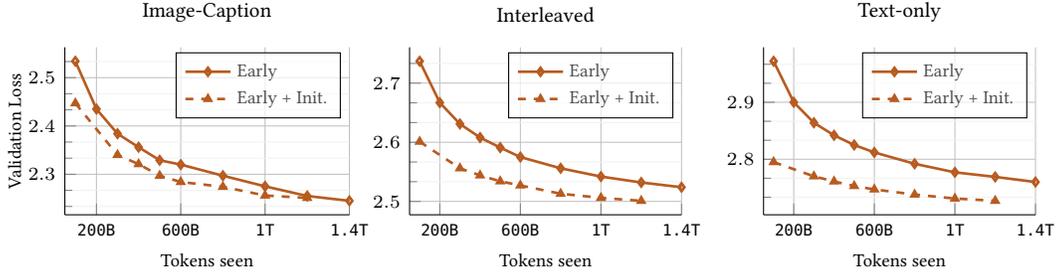

    \centering
    \captionsetup{type=figure}
    \begin{subfigure}[t]{0.34\linewidth}
        \input{graphs/early/early_vs_early_init_scaledata_getty}
    \end{subfigure}
    \hfill
    \begin{subfigure}[t]{0.31\linewidth}
        \input{graphs/early/early_vs_early_init_scaledata_obelics}
    \end{subfigure}
    \hfill
    \begin{subfigure}[t]{0.31\linewidth}
        \input{graphs/early/early_vs_early_init_scaledata_dclm}
    \end{subfigure}
    \vspace{-15pt}
    \begin{center}
        \ref{sharedlegend}
    \end{center}
    \vspace{-4pt}
    \caption{\textbf{Early native vs initializing from LLMs: initializing from
    pre-trained models and scaling training tokens.} We compare training with and
    without initializing from DCLM-1B.}
    \label{fig:early_vs_early_init_scaledata}
\end{figure}

%% file: figs/scaling_laws_moes.tex
\begin{figure}[t!]
    \centering
    \captionsetup{type=figure}
    \begin{subfigure}[t]{1.0\linewidth}
        \input{graphs/moe/moe_scaleflops_avg}
    \end{subfigure}
    \vspace{-7mm}
    \begin{tikzpicture}
        \node[anchor=north] (legend) at (0\linewidth, 0) {
            \begin{axis}[
                        hide axis, %
                        xmin=0, xmax=0.5, ymin=0, ymax=1, %
                        legend columns=1, %
                        legend style={
                            at={(0.25, 0.48)}, %
                            anchor=north, %
                            /tikz/every even column/.append style={column sep=0.2cm}, %
                            scale=0.5,
                            cells={align=left}, font=\scriptsize,
                            anchor=center,
                        },
                    ]
                \addlegendimage{legend moe_0_2b style}
                \addlegendentry{0.275B}
                \addlegendimage{legend moe_0_4b style}
                \addlegendentry{0.464B}
                \addlegendimage{legend moe_0_9b style}
                \addlegendentry{0.932B}
                \addlegendimage{legend moe style}
                \addlegendentry{1.627B}
                \addlegendimage{legend moe_2_2b style}
                \addlegendentry{2.280B}
                \addlegendimage{legend moe_3_3b style}
                \addlegendentry{3.354B}
            \end{axis}
        };
    \end{tikzpicture}
    \vspace{0.2cm}
    \caption{\textbf{Scaling laws for sparse early-fusion NMMs.} We report the final validation loss averaged across interleaved,
    image-captions and text data.}
    \label{fig:early_scaleflops_moe_avg}
\end{figure}

%% file: figs/dense_vs_moe_scaledata.tex
\begin{figure}[t!]
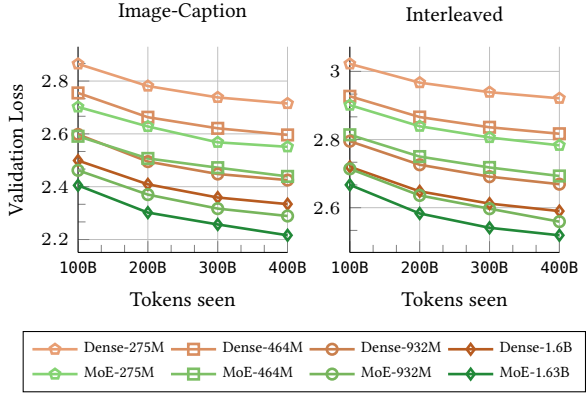

    \centering
    \captionsetup{type=figure}
    \begin{subfigure}[t]{0.49\linewidth}
        \input{graphs/dense_moe/dense_vs_moe_scaledata_getty}
    \end{subfigure}
    \begin{subfigure}[t]{0.49\linewidth}
        \input{graphs/dense_moe/dense_vs_moe_scaledata_obelics}
    \end{subfigure}
    \vspace{-15pt}
    \begin{center}
        \ref{sharedlegend}
    \end{center}
    \caption{\textbf{MoE vs Dense: scaling training FLOPs.} We compare MoE and
    dense early-fusion models when scaling both the amount of training tokens
    and model sizes. MoEs beat dense models when matching the
    number of active parameters.}
    \label{fig:dense_vs_moe_scaledata}
\end{figure}

%% file: tables/sft_results.tex
\begin{table}[h!]
    \centering
    \setlength{\tabcolsep}{3pt} %
    \renewcommand{\arraystretch}{1.3} %
    \resizebox{1\linewidth}{!}{
    \begin{tabular}{lcccccccc}
        & \multicolumn{6}{c}{Accuracy} & \multicolumn{2}{c}{CIDEr}  \\
        \cmidrule(lr){2-7} \cmidrule(lr){8-9}
        & AVG & VQAv2 & TextVQA & OKVQA & GQA & VizWiz & COCO & TextCaps \\
        \shline
        Late-fusion   & 46.8  & 69.4  & 25.8  & 50.1  & \textbf{65.8}  & 22.8  & 70.7  & 50.9 \\
        Early-fusion  & 47.6  & 69.3  & 28.1  & \textbf{52.1}  & 65.4  & 23.2  & \textbf{72.0}  & 53.8 \\
        Early-MoEs    & \textbf{48.2}  & \textbf{69.8}  & \textbf{30.0}  & \textbf{52.1}  & 65.4  & \textbf{23.6}  & 69.6  & \textbf{55.7} \\
    \end{tabular}%
    }
    \caption{\textbf{Supervised finetuning on the LLaVA mixture.} All models are native at 1.5B scale and pre-trained on 300B tokens.}
    \vspace{-7pt}
    \label{tab:sft}
\end{table}

%% file: figs/soft_vs_hard_routing.tex
\begin{figure}[t!]
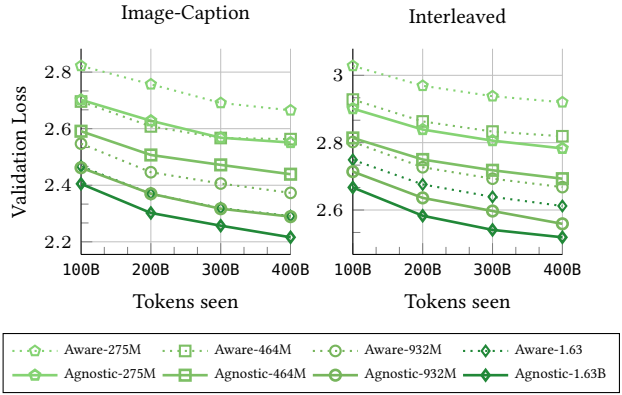

    \centering
    \captionsetup{type=figure}
    \begin{subfigure}[t]{0.49\linewidth}
        \input{graphs/hard_moe/hard_vs_moe_scaledata_getty}
    \end{subfigure}
    \begin{subfigure}[t]{0.49\linewidth}
        \input{graphs/hard_moe/hard_vs_moe_scaledata_obelics}
    \end{subfigure}
    \vspace{-15pt}
    \begin{center}
        \ref{sharedlegend}
    \end{center}

    \caption{\textbf{Modality-aware vs modality agnostic routing for sparse NMMs.} We compare
    modality-agnostic routing with modality-aware routing
    when scaling both the amount of training tokens and model sizes.}
    \label{fig:hard_vs_moe_scaledata}
\end{figure}

%% file: sec/2_related_work.tex
\section{Related work}

\cpar{Large multimodal models.} A long-standing research goal has been to develop models capable of perceiving the world through multiple modalities, akin to human sensory experience.  Recent progress in vision and language processing has shifted the research focus from smaller, task-specific models toward large, generalist models that can handle diverse inputs \cite{team2023gemini,hurst2024gpt4o}.  Crucially, pre-trained vision and language backbones often require surprisingly little adaptation to enable effective cross-modal communication \cite{tsimpoukelli2021multimodalfrozen,shukor2023epalm,vallaeys2024improveddepalm,merullo2023linearly,koh2023grounding}.  Simply integrating a vision encoder with either an encoder-decoder architecture \cite{shukor2023unival,wang2022ofa,lu2022unified,mizrahi20234m} or a decoder-only LLM has yielded highly capable multimodal systems \cite{laurenccon2024mattersidefics2,alayrac2022flamingo,liu2024improvedllava,wang2024qwen2,xue2024xgenblip3,chen2024internvl,zhu2024minigpt,abdin2024phi3,dai2024nvlm,beyer2024paligemma,moon2024anymal,shukor2025smolvla}. This late-fusion approach, where modalities are processed separately before being combined, is now well-understood, with established best practices for training effective models \cite{laurenccon2024obelics,mckinzie2025mm1,zhang2024mm1_5,lin2024vila}.  In contrast, early-fusion models \cite{fuyu8b,team2024chameleon,diao2024unveiling}, which combine modalities at an earlier stage, remain relatively unexplored, with only a limited number of publicly released models \cite{fuyu8b,diao2024unveiling}.  Unlike \cite{diao2024unveiling,team2024chameleon}, our models utilize only a single linear layer and rely exclusively on a next-token prediction loss. Furthermore, we train our models from scratch on all modalities without image tokenization.

\input{figs/moe_entropy}

\cpar{Native Multimodal Models.} We define native multimodal models as those trained from scratch on all modalities simultaneously \cite{team2023gemini} rather than adapting LLMs to accommodate additional modalities. Due to the high cost of training such models, they remain relatively underexplored, with most relying on late-fusion architectures \cite{kosmoshuang2023language,yu2022coca}. Some multimodal models trained from scratch \cite{aghajanyan2022cm3,team2024chameleon,wang2024emu3} relax this constraint by utilizing pre-trained image tokenizers such as \cite{vqgan,vqvae} to convert images into discrete tokens, integrating them into the text vocabulary. This approach enables models to understand and generate text and images, facilitating a more seamless multimodal learning process.

\cpar{Scaling laws.} Scaling law studies aim to predict how model
performance scales with training compute. Early works
\cite{kaplan2020scaling,hoffmann2022training} found that LLM performance follows
a power-law relationship with compute, enabling the compute-optimal estimation of the number of model parameters and training tokens at scale for a given budget. Similar research has
extended these findings to sparse Mixture of Experts (MoE) models, considering
factors such as sparsity, number of experts, and routing granularity
\cite{krajewski2024scalingmoe,clark2022unifiedscalingmoe,wangscalingmoe}.
Scaling laws have also been observed across various domains, including image
models \cite{fini2024multimodalaimv2}, video models
\cite{rajasegaran2025empirical}, protein LLMs \cite{scalingprotein}, and
imitation learning \cite{pearce2024scaling}. However, few studies have
investigated scaling laws for multimodal models.
Notably,~\citet{aghajanyan2023scalingmm} examined multimodal models that tokenize
modalities into discrete tokens and include multimodal generation. In contrast,
we focus on studying early-fusion models that take raw multimodal inputs and
are trained on interleaved multimodal data.

\cpar{Mixture of experts (MoEs).} MoEs~\cite{shazeer2017outrageously} scale model capacity efficiently by sparsely activating parameters, enabling large models with reduced per-sample compute. While widely studied in LLMs~\cite{fedus2022switch,sun2024hunyuan,jiang2024mixtral,liu2024deepseekv3,wei2024skywork,zoph2022st,lewis2021base,lepikhin2020gshard}, MoEs remain underexplored in multimodal settings. Prior work has examined contrastive models~\cite{mustafa2022multimodal}, late-fusion LLMs~\cite{lin2024moe,li2024aria}, and modality-specific experts~\cite{bao2021vlmo,chen2024eve,shen2023scaling}. We focus on analyzing MoEs in early-fusion multimodal models.

\input{figs/moe_specialization}

%% file: figs/moe_entropy.tex
\begin{figure}[t!]
\centering
\captionsetup{type=figure}

\begin{tikzpicture}
\begin{axis}[
    grid=major, %
    grid style={line width=.1pt, draw=gray!30}, %
    major grid style={line width=.2pt,draw=gray!50},
    minor tick num=2,
    axis x line*=bottom,
    axis y line*=left,
    xmin=0,
    xmax=23,
    ymin=0.0,
    ymax=1.0,
    height=1.6in,
    width=1.0\linewidth,
    ylabel style={align=center, font=\footnotesize},
    xlabel style={font=\footnotesize},
    ylabel=\footnotesize{I/T specialization},
    xlabel={\footnotesize{Layers}},
    ytick distance=0.2,
    yticklabel style={font=\footnotesize, /pgf/number format/fixed, /pgf/number format/precision=2},
    xticklabel style={font=\footnotesize},
    xtick={0,2,4,6,8,10,12,14,16,18,20,22},
    xticklabels={0,2,4,6,8,10,12,14,16,18,20,22},
    mark options={solid},
    legend style={cells={align=left}, font=\scriptsize, text=black}, %
    legend columns=4, %
    legend cell align={left},
    legend to name=sharedlegend,
]

\addplot[color=blue!40!gray, mark=*, mark size=1.7pt, line width=1pt] plot coordinates {  %
(0, 0.9769479348349329)
(1, 0.7427025603097698)
(2, 0.6762374582380828)
(3, 0.6340138224507533)
(4, 0.5783047019889138)
(5, 0.6195673358863271)
(6, 0.5737861572855565)
(7, 0.7066104505575633)
(8, 0.6446960774170309)
(9, 0.6386074080824291)
(10, 0.6128197724774395)
(11, 0.620867456928868)
(12, 0.6240480502092688)
(13, 0.6712385178771478)
(14, 0.591805441066985)
(15, 0.6531511085896324)
(16, 0.6873617200378165)
(17, 0.7072043647262733)
(18, 0.6290614078343542)
(19, 0.6162376387022552)
(20, 0.6237776803436119)
(21, 0.649774290665718)
(22, 0.6745086003424091)
(23, 0.810597818982423)
};
\addlegendentry{HQITP}

\addplot[color=red!40!gray, mark=square*, mark size=1.7pt, line width=1pt] plot coordinates { %
(0, 0.8098464848229123)
(1, 0.6803114998878562)
(2, 0.640428138604904)
(3, 0.15356616480436047)
(4, 0.13402442087930844)
(5, 0.33860156581505885)
(6, 0.2894806232076399)
(7, 0.269278406260023)
(8, 0.17195894943324286)
(9, 0.15091557473173656)
(10, 0.2050356661788283)
(11, 0.21603190628094793)
(12, 0.22923064654990588)
(13, 0.34291773221125066)
(14, 0.15175535984919586)
(15, 0.16059145647036643)
(16, 0.12438172525723101)
(17, 0.20722842981400202)
(18, 0.12145399323126793)
(19, 0.12335098001817868)
(20, 0.1811717532056517)
(21, 0.13129565163316004)
(22, 0.19701360887683683)
(23, 0.28035775702112686)
};
\addlegendentry{Obelics}
\end{axis}

\end{tikzpicture}
\vspace{-10pt}
\begin{center}
    \ref{sharedlegend}
\end{center}

\caption{\textbf{MoE specialization score.} Entropy-based image/text specialization score (as described in~\cref{sec:specialization}) across layers for two data sources: HQITP and Obelics. HQITP has a more imbalanced image-to-text token distribution, resulting in generally higher specialization. Despite this difference, both data sources exhibit a similar trend: the specialization score decreases in the early layers before increasing again in the final layers.
}
\label{fig:tokens_specialization}
\end{figure}

%% file: figs/moe_specialization.tex
\begin{figure}[t!]
    \centering
    \captionsetup{type=figure}
    \begin{subfigure}[h]{1\linewidth}
    \includegraphics[height=0.28\textwidth]{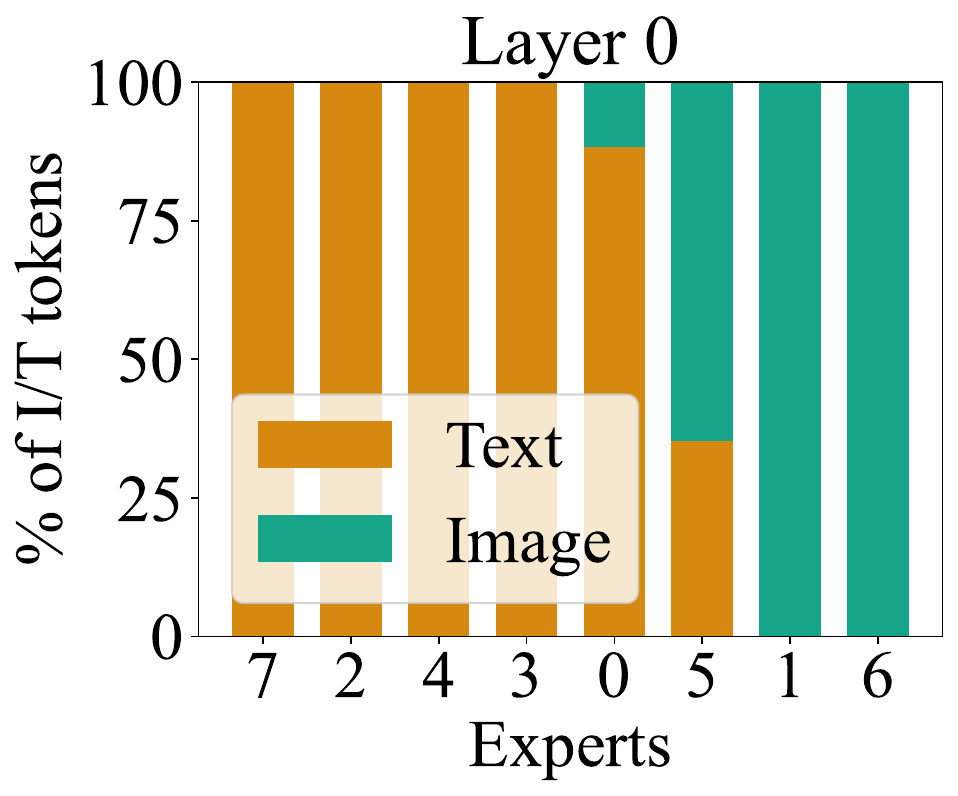}
    \includegraphics[height=0.28\textwidth]{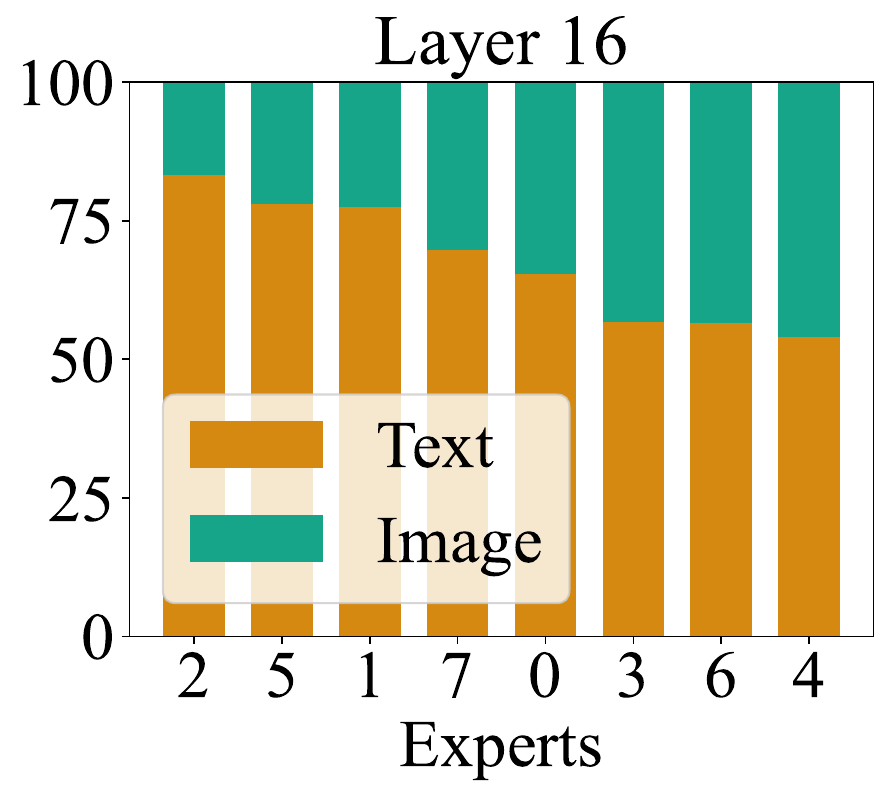}
    \includegraphics[height=0.28\textwidth]{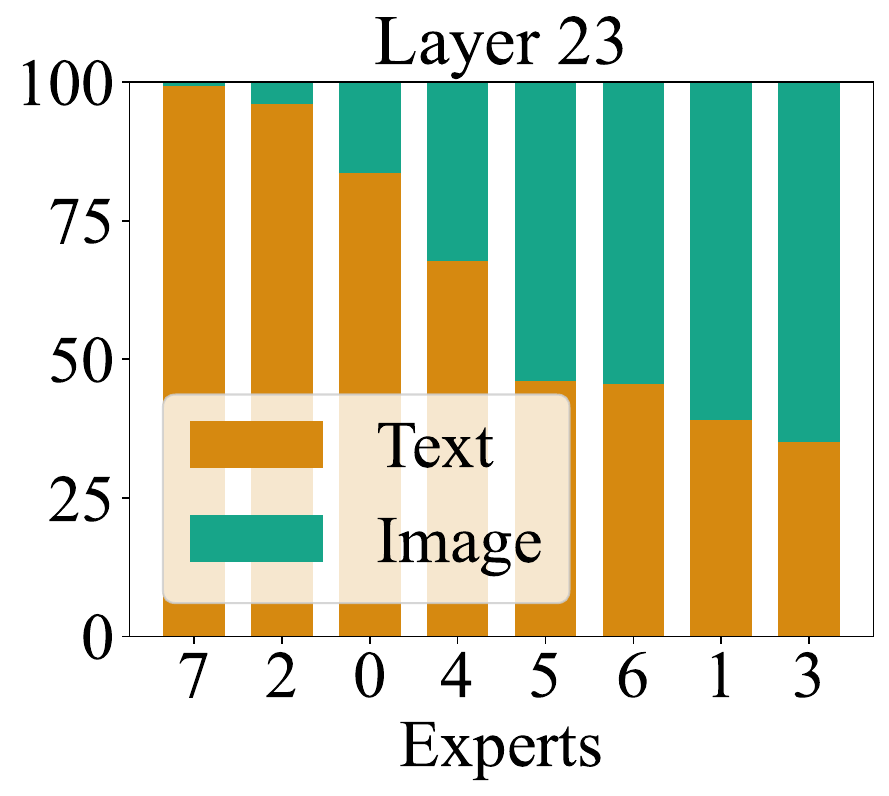}
    \end{subfigure}    \caption{\textbf{MoE specialization frequency.} Percentage of text and image tokens routed to each expert on interleaved data from Obelics. Experts are ordered for better visualization. The first layer shows the highest amount of unimodal experts.}
    \label{fig:tokens_assignment}
\end{figure}

%% file: sec/3_conclusion.tex
\section{Limitations} 

Our study finds that scaling law coefficients are broadly consistent across training mixtures, though a broader exploration is needed to validate this observation. While validation loss scales predictably with compute, the extent to which this correlates with downstream performance remains unclear and warrants further investigation. The accuracy of scaling law predictions improves with higher FLOPs, but their extrapolation to extreme model sizes is still an open question (\Cref{app:limitation} for more details).

\section{Conclusion} 
We explore various strategies for compute-optimal pretraining of native multimodal models. We found the NMMs follow similar scaling laws to those of LLMs. Contrary to common belief, we find no inherent advantage in adopting late-fusion architectures over early-fusion ones. While both architectures exhibit similar scaling properties, early-fusion models are more efficient to train and outperform late-fusion models at lower compute budgets. Furthermore, we show that sparse architectures encourage modality-specific specialization, leading to performance improvements while maintaining the same inference cost.

\section*{Acknowledgment} We thank Philipp Dufter, Samira Abnar, Xiujun
Li, Zhe Gan, Alexander Toshev, Yinfei Yang, Dan Busbridge, and Jason Ramapuram
for many fruitful discussions. We thank Denise Hui, and Samy Bengio for infra
and compute support. Finally, we thank, Louis Béthune, Pierre Ablin, Marco
Cuturi, and the MLR team at Apple for their support throughout the project.

%% file: sec/X_suppl.tex
\clearpage
\maketitlesupplementary
\appendix

\noindent This supplementary material is organized as follows:
\vspace{0.3cm}
\begin{itemize}
    \item \Cref{app:implementation_details}: contains the implementation details and the hyperparameters used to train our models.
    \item \Cref{app:late_vs_early}: contains detailed comparison between early and late fusion  models.
    \item \Cref{app:scaling_laws}: contains more details about scaling laws derivation, evaluation and additional results.
    \item \Cref{app:limitation}: contains discussion about the paper limitations.
    \item \Cref{app:moes}: contains more results about MoEs and modality specialization.
\end{itemize}

\section{Experimental setup}
\label{app:implementation_details}

In \Cref{tab:scaling_laws_hparams}, we show the pre-training hyperparameters for different model configurations used to derive the scaling laws. The number of parameters ranges from 275M to 3.7B, with model width increasing accordingly, while the depth remains fixed at 24 layers. Learning rates vary by model size, decreasing as the model scales up. Based on empirical experiments and estimates similar to \cite{mckinzie2025mm1}, we found these values to be effective in our setup. Training is optimized using a fully decoupled AdamW optimizer with momentum values $\beta_1=0.9$, $\beta_2=0.95$, and a weight decay of $1\text{e}{-4}$. The batch size is set to 2k samples, which account for 2M tokens, given 1k context length.  Gradient clipping is set to 1.0, with a maximum warmup duration of 5k iterations, adjusted for shorter training runs: 1k and 2.5k warmup steps for models trained between 1k–4k and 5k–15k steps, respectively. For MoEs, we found that longer warmup is significantly better, so we adopt a 2.5k warmup for all runs under 20k steps. We use a constant learning rate schedule with cooldown during the final 20\% of training, gradually reducing to zero following an inverse square root schedule. For vision processing, image inputs are divided into $(14,14)$ patches, with augmentations including Random Resized Crop (resizing images to 224px with a scale range of [0.4, 1.0]) and Random Horizontal Flip with a probability of 0.5.  We train our models on mixture of interleaved, image captions and text only data \Cref{tab:pretraining_datasets}.
For late fusion models, we found that  using smaller learning rate for the vision encoder significantly boost the performance \Cref{tab:late_scaler_scratch}, and when both the encoder and decoder are initialized (\Cref{sec:app_init_early_late}) we found that freezing the vision encoder works best \Cref{tab:late_scaler_init}.

 \begin{table}[h]
    \centering
    \setlength{\tabcolsep}{12pt}
    \renewcommand{\arraystretch}{1}
    \resizebox{1\linewidth}{!}{
    \begin{tabular}{lccrc}
        Data type & dataset & \#samples & sampling prob. \\
         \shline
         \multirow{3}{*}{Image-Caption} &  DFN~\citep{fang2023data} &
         2B & 27\% \\
          & COYO~{\citep{kakaobrain2022coyo700m}} &  600M & 11.25\% \\
          & HQITP\citep{ranasinghe2023perceptual}  & 400M & 6.75\% \\
          Interleaved & Obelics \cite{laurenccon2024obelics}  & 141M Docs &
          45\% \\
          Text & DCLM \cite{li2024datacomp} & 6.6T Toks & 10\% \\
          
    \end{tabular}} 
    \caption{\textbf{Pre-training data mixture.} Unless otherwise specified, the training mixture contains 45\%, 45\% and 10\% of image captions, interleaved documents and text-only data.}
    \label{tab:pretraining_datasets}
    \vspace{-5pt}
\end{table}

\begin{table}[htb]
    \begin{center}
        \centering
        \setlength{\tabcolsep}{4pt}
        \resizebox{\linewidth}{!}{
        \begin{tabular}{l c c c c c c}
            \toprule
            \textbf{Early-fusion} \\
            \midrule
            Params &  275M & 468M & 932M  & 1.63B & 2.28B & 3.35B \\
            width & 800 & 1088 & 1632 & 2208 & 2624 & 3232\\
            depth & \multicolumn{6}{c}{24} \\
            Learning rate & 1.5e-3 & 1.5e-3 & 5e-4 & 4.2e-4 & 4e-4 & 3.5e-4 \\
            \midrule
            \textbf{Late-fusion} \\
            \midrule
            Params &  289M & 494M & 1B  & 1.75B & 2.43B & 3.7B \\
            vision encoder width & 384 & 512 & 768 & 1024 & 1184 & 1536 \\
            vision encoder depth & \multicolumn{6}{c}{24} \\
            width & 768 & 1024 & 1536 & 2048 & 2464 & 3072\\
            depth & \multicolumn{6}{c}{24} \\
            Learning rate & 1.5e-3 & 1.5e-3 & 5e-4 & 4.2e-4 & 3.8e-4 & 3.3e-4 \\
            \midrule
            \textbf{Early-fusion MoEs} \\
            \midrule
            Active Params &  275M & 468M & 932M  & 1.63B & 2.28B & 3.35B \\
            width & 800 & 1088 & 1632 & 2208 & 2624 & 3232\\
            depth & \multicolumn{6}{c}{24} \\
            Learning rate & 1.5e-3 & 1.5e-3 & 5e-4 & 4.2e-4 & 4e-4 & 3.5e-4 \\
            \midrule
            Training tokens & \multicolumn{6}{c}{2.5B-600B} \\
            Optimizer & \multicolumn{6}{c}{Fully decoupled AdamW~\cite{loshchilov2017decoupled}} \\ %
            Optimizer Momentum & \multicolumn{6}{c}{$\beta_1=0.9 ,\beta_2=0.95$} \\
            Minimum Learning rate & \multicolumn{6}{c}{0} \\
            Weight decay & \multicolumn{6}{c}{1e-4} \\
            Batch size & \multicolumn{6}{c}{2k} \\
            Patch size & \multicolumn{6}{c}{(14, 14)} \\
            Gradient clipping & \multicolumn{6}{c}{1.0} \\
            MAximum Warmup iterations & \multicolumn{6}{c}{5k} \\
            Augmentations: \\
            \quad {\tt RandomResizedCrop} \\
            \qquad {\tt size} & \multicolumn{6}{c}{224px} \\
            \qquad {\tt scale} & \multicolumn{6}{c}{[0.4, 1.0]} \\
            \quad {\tt RandomHorizontalFlip} & \multicolumn{6}{c}{$p=0.5$} \\
            \bottomrule
        \end{tabular}}
    \end{center}
    \caption{\textbf{Pre-training hyperparameters} We detail the hyperaparmeters used for pre-training different model configurations to derive scaling laws.}
    \label{tab:scaling_laws_hparams}
    \end{table}

\begin{table}[htb]
    \centering
    \setlength{\tabcolsep}{4pt}
    \renewcommand{\arraystretch}{1.3}
    \resizebox{0.9\linewidth}{!}{
    \begin{tabular}{lccccc}
        Vision encoder & Interleaved & Image-Caption & Text  & AVG & AVG (SFT)  \\
        lr scaler & (CE) & (CE) & (CE) & (CE) & (Acc) \\
        \shline
        1 & 2.521 & 2.15 & 2.867 &  2.513 & 43.49 \\
        0.1 & 2.502 & 2.066 & 2.862 &  2.477 & 52.27\\
        0.01 & 2.502 & 2.066 & 2.859 &  2.476 & 53.76\\
        0.001 & 2.513 & 2.066 & 2.857 &  2.479 & -- \\
        0 (frozen) & 2.504 & 2.061 & 2.856 & 2.474 & 54.14 \\
        \bottomrule
    \end{tabular}%
    } 
    \caption{\textbf{Vision encoder scaler.} Freezing the vision encoder works best when initializing late-fusion models with pre-trained models.}
    \label{tab:late_scaler_init}
\end{table}

\begin{table}[htb]
    \centering
    \setlength{\tabcolsep}{4pt}
    \renewcommand{\arraystretch}{1.3}
    \resizebox{0.9\linewidth}{!}{
    \begin{tabular}{lccccc}
        Vision encoder & Interleaved & Image-Caption & Text  & AVG & AVG (SFT)  \\
        lr scaler & (CE) & (CE) & (CE) & (CE) & (Acc) \\
        \shline
        0.1 & 2.674 & 2.219 & 3.072   & 2.655 & 34.84 \\
        0.01 & 2.672 & 2.197 & 3.071  & 2.647 & 38.77 \\
        0.001 & 2.674 & 2.218 & 3.073 & 2.655 & 38.46 \\
        \bottomrule
    \end{tabular}%
    } 
    \caption{\textbf{Vision encoder scaler.} Reducing the learning rate for the vision encoder is better when training late-fusion models from scratch.}
    \label{tab:late_scaler_scratch}
\end{table}

\input{figs/late_vs_early_equal_tokens}

\begin{figure*}[htp]
    \centering
    \captionsetup{type=figure}
    \begin{subfigure}[t]{0.33\linewidth}
        \input{graphs/early_late/early_late_datatype_sameflops_obelics}
    \end{subfigure}
    \begin{subfigure}[t]{0.33\linewidth}
         \input{graphs/early_late/early_late_datatype_sameflops_getty}
    \end{subfigure}
    \begin{subfigure}[t]{0.33\linewidth}
         \input{graphs/early_late/early_vs_late_datatype_sameflops_dclm}
    \end{subfigure}
            
    \vspace{0.3cm}
    \caption{\textbf{Early vs late fusion: changing the training mixture.} We vary the training mixtures and plot the final training loss. Early fusion models become better when increasing the proportion of interleaved documents. Early and late fusion has 1.63B and 1.75B parameters respectively.}
    \label{fig:early_vs_late_datatype_sameflops}
\end{figure*}

\section{Late vs early fusion}
\label{app:late_vs_early}
This section provides additional comparison between early and late fusion models.

\subsection{Scaling FLOPs} \Cref{fig:early_vs_late_scaledata_main} compares early-fusion and late-fusion models when scaling FLOPs. Specifically, for each model size, we train multiple models using different amounts of training tokens. The performance gap between the two approaches mainly decreases due to increasing model sizes rather than increasing the number of training tokens. Despite the decreasing gap, across all the models that we train, early-fusion consistently outperform late-fusion.

\subsection{Changing the training data mixture} We analyze how the performance gap between early and late fusion models changes with variations in the training data mixture. As shown in \Cref{fig:early_vs_late_textratio} and \Cref{fig:early_vs_late_datatype_sameflops}, when fixing the model size, increasing the ratio of text and interleaved data favors early fusion. Interestingly, the gap remains largely unchanged for other data types. We also observe interference effects between different data types. Specifically, increasing the amount of interleaved data negatively impacts performance on image captions and vice versa. Additionally, increasing the proportion of text-only data slightly improves interleaved performance but increases loss on image captions. Overall, we find that text-only and interleaved data are correlated across different setups.

\begin{figure*}[htp]
    \centering
    \captionsetup{type=figure}
    \begin{subfigure}[t]{0.33\linewidth}
        \input{graphs/early_late/early_vs_late_textratio_obelics}
    \end{subfigure}
    \begin{subfigure}[t]{0.33\linewidth}
        \input{graphs/early_late/early_vs_late_textratio_getty}
    \end{subfigure}
    \begin{subfigure}[t]{0.33\linewidth}
        \input{graphs/early_late/early_vs_late_textratio_dclm}
    \end{subfigure}
            
    \vspace{0.3cm}
    \caption{\textbf{Early vs late fusion: changing the amount of text-only data in the training mixture (isoFLOPs).} We vary the ratio of text-only data and plot the final training loss. The gap increases with the text data ratio in favor of early fusion model. Early fusion has 1.63B parameters and late fusion 1.75B parameters.}
    \label{fig:early_vs_late_textratio}
\end{figure*}

\input{figs/early_vs_late_imageres}

\subsection{Scaling image resolution is in favor of early-fusion}

We examine how both
architectures perform with varying image resolution. We fix the number of model parameters to 1.63B and 1.75B for early and late fusion respecively. All models are trained for 100K steps or 200B tokens. Since the patch size remains
constant, increasing the resolution results in a higher number of visual tokens. For all resolutions, we maintain the same number of text tokens.
As shown in \Cref{fig:early_vs_late_imageres}, the early-fusion model
consistently outperforms the late-fusion model across resolutions, particularly
for multimodal data, with the performance gap widening at higher resolutions.
Additionally, we observe that the loss on text and interleaved data increases as
resolution increases.

\subsection{Early-fusion is consistently better when matching the late-fusion model size}
\input{figs/early_vs_late_datatype_isoparams}

In this section, we compare the late-fusion model with different configurations
of early-fusion one. Specifically, we train early-fusion models that match the
late-fusion model in total parameters (Params), text model size (Text), and
FLOPs (FLOPs), assuming 45-45-10 training mixture. As shown in
\Cref{fig:early_vs_late_datatype_isoparams}, early fusion consistently
outperforms late fusion when normalized by total parameters, followed by
normalization by FLOPs. When matching the text model size, early fusion performs
better at higher ratios of interleaved data.

\subsection{Different late-fusion configuration} 
We examine how this scaling changes with different late-fusion configurations. Instead of scaling both the vision and text models equally, as done in the main paper, we fix the vision encoder size to 300M and scale only the text model. \Cref{fig:early_vs_late_scalellmdata_dclm} shows that late-fusion models lag behind at smaller model sizes, with the gap closing significantly as the text model scales. This suggests that allocating more parameters to shared components is more beneficial, further supporting the choice of early-fusion models.

\begin{figure*}[htp]
    \centering
    \captionsetup{type=figure}
    \begin{subfigure}[t]{0.33\linewidth}
        \input{graphs/early_late/early_vs_late_scalellmdata_getty}
    \end{subfigure}
    \begin{subfigure}[t]{0.33\linewidth}
        \input{graphs/early_late/early_vs_late_scalellmdata_obelics}
    \end{subfigure}
    \begin{subfigure}[t]{0.33\linewidth}
        \input{graphs/early_late/early_vs_late_scalellmdata_dclm}
    \end{subfigure}

    \makebox[0.9\linewidth]{ %
        \begin{tikzpicture}
            \begin{axis}[
                hide axis, %
                xmin=0, xmax=0.5, ymin=0, ymax=1, %
                legend columns=4, %
                legend style={
                    at={(0.5, 1)}, %
                    anchor=north, %
                    /tikz/every even column/.append style={column sep=0.2cm}, %
                    scale=0.5, %
                    cells={align=left}, font=\footnotesize,
                },
            ]
               \addlegendimage{legend late_0_2b style}
                \addlegendentry{Late-0.555B}
                \addlegendimage{legend late_0_4b style}
                \addlegendentry{Late-1.14B}
                \addlegendimage{LateGradStart!50!LateGradEnd, thick, solid, mark=*, mark size=1.5pt}
                \addlegendentry{Late-2.320B}
                \addlegendimage{LateGradStart!75!LateGradEnd, thick, solid, mark=*, mark size=1.5pt}
                \addlegendentry{Late-3.33B}

                \addlegendimage{legend early_0_2b style}
                \addlegendentry{Early-0.464B}
                \addlegendimage{EarlyGradStart!25!EarlyGradEnd, thick, solid, mark=*, mark size=1.5pt}
                \addlegendentry{Early-0.932B}
                \addlegendimage{legend early style}
                \addlegendentry{Early-1.627B}
                \addlegendimage{legend early_2_2b style}
                \addlegendentry{Early-3.354B}
            \end{axis}
        \end{tikzpicture}
    }

    \vspace{-4.2cm}
    \caption{\textbf{Early vs late fusion: scaling training FLOPs while fixing the vision encoder size.} We compare early and late fusion models when scaling both the amount of training tokens and model sizes. For late fusion mdoels, we fix the vision encoder size (300M) and scale the text model (250M, 834M, 2B, 3B). The gap between early and late get tighter when scaling the text model.}
    \label{fig:early_vs_late_scalellmdata_dclm}
\end{figure*}

\subsection{Different context lengths}

\begin{figure}[!h]
    \centering
    \begin{subfigure}[t]{0.7\linewidth}
        \input{graphs/early_late/early_vs_late_scalecontext_obelics}
    \end{subfigure}
    \caption{\textbf{Early vs late fusion with different context lengths.}} %
    \label{fig:early_late_context}
\end{figure}

 In the paper, we use a 1k context length following \citep{kaplan2020scaling}. Also following, this paper, we ignore the context length effect, as the model dimension dominates the training compute estimate. Moreover, \cite{pearce2024reconciling} empirically found that scaling coefficients are robust to context length. Nevertheless, Our initial experiments (\Cref{fig:early_late_context}) indicate that scaling the context length did not significantly affect the comparison between late and early fusion.

\subsection{Initializing from LLM and CLIP}
\label{sec:app_init_early_late}

We study the case where both late and early fusion models are initialized from pre-trained models, specifically DCLM-1B \cite{li2024datacomp} and CLIP-ViT-L \cite{radford2021learning} for late fusion. Interestingly, \Cref{fig:early_vs_late_init_scaledata} shows that for text and interleaved multimodal documents, early fusion can match the performance of late fusion when trained for longer. However, closing the gap on image caption data remains more challenging. Notably, when considering the overall training cost, including that of pre-trained models, early fusion requires significantly longer training to compensate for the vision encoder’s pretraining cost.

\input{figs/early_vs_late_init_scaledata}

\section{Scaling laws}
\label{app:scaling_laws}

\subsection{Fitting \(L = F(N, D)\)}  
Following \cite{hoffmann2022training}, we determine the parameters that minimize the following objective across all our runs \(i\):  
\begin{equation}
\footnotesize
    \min_{a,b,e,\alpha,\beta} \sum_{i} \text{Huber}_\delta \left( \text{LSE} \left( a - \alpha \log N_i, b - \beta \log D_i, e \right) - \log L_i \right),
\end{equation}  
We perform this optimization across various initialization ranges and select the parameters that achieve the lowest loss across all initializations. Specifically, our grid search spans \(\{0, 0.5, 2.5\}\) for \(\alpha\) and \(\beta\), \(\{0, 5, 10, ..., 30\}\) for \(a\) and \(b\), and \(\{-1, -0.5, 1, 0.5\}\) for \(e\). We use the L-BFGS algorithm with \(\delta=1e-3\).

\subsection{Fitting \(N \propto C^a\), \(D \propto C^b\), \(D \propto N^d\)}  
While these equations have a closed-form solution \cite{hoffmann2022training} for early-fusion models that can be derived from \Cref{eq:scaling_laws}, this is not the case for late-fusion models without specifying either the vision encoder or text model size. To ensure a fair comparison, we derive these equations for both models, by performing linear regression in log space. We found that the regression is very close to the coefficient found with closed-form derivation \Cref{tab:scaling_laws_closed_form}. For instance, to derive \(N = K_aC^a\), given a FLOP budget \(C\) and a set of linearly spaced tokens \(D_i\) ranging from 10B to 600B, we compute the model size for each \(D_i\) as \(N_i = \frac{C}{6D}\) for early fusion and \(N_i = \frac{C}{6D}+0.483*N_v\) for late fusion (for the 45-45-10 mixture, \(D_v=0.544D\), thus $C=6D(0.544N_v+N_t)$). We then apply \Cref{eq:scaling_laws} to obtain the loss for each model size and select \(N\) that has the minimum loss. We repeat this for all FLOP values corresponding to our runs, resulting in a set of points \((C, N_{opt})\) that we use to regress \(a\) and \(K_a\).  We follow a similar procedure to find \(b\) and \(d\). For late-fusion models, we regress a linear model to determine \(N_v\) given \(N\). Notably, even though we maintain a fixed width ratio for late-fusion models, this approach is more accurate, as embedding layers prevent a strictly fixed ratio between text and vision model sizes. We present the regression results in \Cref{fig:scaling_laws_closed_form_early_late}.

\begin{table}[htb]
    \centering
    \setlength{\tabcolsep}{6pt} %
    \renewcommand{\arraystretch}{1.3} %
    \resizebox{0.9\linewidth}{!}{
    \begin{tabular}{lcccccccc}
        Model & $a$ & $b$ & $d$ & $n$ & $dn$  \\
        \midrule
        Closed form  & 0.52649 & 0.47351 & 0.89938 &  1.11188 & -0.05298  \\
        Regression & 0.52391 & 0.47534 & 0.90052 & 1.10224 & -0.04933  \\
        \bottomrule
    \end{tabular}%
    }
    \caption{\textbf{Scaling laws parameters for early-fusion.} Doing regression to derive the scaling laws coefficients leads to very close results to using the closed-form solution.}
    \label{tab:scaling_laws_closed_form}
\end{table}

\subsection{Fitting \(L \propto C^c\)}  
To determine the relationship between the final model loss and the compute budget \(C\), we begin by interpolating the points corresponding to the same model size and compute the convex hull that covers the minimum loss achieved by all runs for each FLOP. This results in a continuous mapping from the FLOPs to the lowest loss. We consider a range of FLOPs, excluding very small values ($\leq 3e^{19}$), and construct a dataset of \((C, L)\) for linearly spaced compute \(C\). Using this data, we find the linear relationship between \(L\) and \(C\) in the log space and deduce the exponent \(c\). We visualize the results in \Cref{fig:scaling_laws_early_late_moe}.

\begin{figure}[h!]
    \centering
    \captionsetup{type=figure}
    \begin{subfigure}[t]{1\linewidth}

    \begin{subfigure}[t]{0.49\linewidth}
        \input{graphs/early/early_scalinglaws_params_vs_flops_avg_ap3}
    \end{subfigure}
    \begin{subfigure}[t]{0.49\linewidth}
        \input{graphs/late/late_scalinglaws_params_vs_flops_avg_ap3}
    \end{subfigure}
    
    \begin{subfigure}[t]{0.49\linewidth}
        \input{graphs/early/early_scalinglaws_tokens_vs_flops_avg_ap3}
    \end{subfigure}
    \begin{subfigure}[t]{0.49\linewidth}
        \input{graphs/late/late_scalinglaws_tokens_vs_flops_avg_ap3}
    \end{subfigure}
    
    \begin{subfigure}[t]{0.49\linewidth}
        \input{graphs/early/early_scalinglaws_tokens_to_params_vs_flops_avg_ap3}
    \end{subfigure}
    \begin{subfigure}[t]{0.49\linewidth}
        \input{graphs/late/late_scalinglaws_tokens_to_params_vs_flops_avg_ap3}
    \end{subfigure}

    \end{subfigure}
    \caption{\textbf{Regression results of the scaling laws coefficients.} our estimation of the scaling coefficients is close to the closed form solution.}
    \label{fig:scaling_laws_closed_form_early_late}
\end{figure}

\begin{figure*}[h!]
    \centering
    \captionsetup{type=figure}
    \begin{subfigure}[t]{0.43\linewidth}
        \input{graphs/early_late/pred_loss_vs_loss_early}
    \end{subfigure}\hspace{0.6cm}
    \begin{subfigure}[t]{0.43\linewidth}
        \input{graphs/late/pred_loss_vs_loss_late}
    \end{subfigure}
    
    \caption{\textbf{Observed vs predicted loss.} We visualize the loss predicted by our scaling laws (\Cref{eq:scaling_laws}) and the actual loss achived by each run.}
    \label{fig:observed_vs_predicted_loss}
\end{figure*}

\subsection{Scaling laws for different target data type}
In \Cref{fig:scaling_laws_early_late_moe_getty_obelics_dclm}, we derive the scaling laws for different target data types. In general, we observe that the model learns image captioning faster than interleaved data, as indicated by the higher absolute value of the scaling exponent (e.g., 0.062 vs 0.046), despite using the same data ratio for captioning and interleaved data (45\% each). Additionally, we find that the model learns more slowly on text-only data, likely due to the smaller amount of text-only data (10\%). Across model configurations, we find that early fusion scales similarly to late fusion on image captioning but has a lower multiplicative constant (49.99 vs 47.97). For MoEs, the model learns faster but exhibits a higher multiplicative constant. On text and interleaved data, early and late fusion models scale similarly and achieve comparable performance. However, MoEs demonstrate better overall performance while learning slightly more slowly.

\subsection{Scaling laws for different training mixtures}

We investigate how the scaling laws change when modifying the training mixtures. Specifically, we vary the ratio of image caption, interleaved, and text-only data and report the results in \Cref{fig:app_early_scaleflops_data_mixtures}. Overall, we observe similar scaling trends, with only minor changes in the scaling coefficients. Upon closer analysis, we find that increasing the ratio of a particular data type in the training mixture, leads to a corresponding increase in its scaling exponent. For instance, increasing the ratio of image captions from 30\% to 40\% raises the absolute value of the exponent from 0.056 to 0.061. However, for text-only data, we do not observe significant changes in the scaling coefficients when varying its proportion in the training mixture.

\begin{table}[htb]
    \centering
    \setlength{\tabcolsep}{6pt} %
    \renewcommand{\arraystretch}{1.3} %
    \resizebox{0.7\linewidth}{!}{
    \begin{tabular}{lccc}
        Parameter & MSE & R2 & MAE (\%) \\
        \shline
        Held-in      & 0.0029  & 0.9807 & 0.8608 \\
        Held-out     & 0.0004 & 0.9682 & 0.5530 \\
    \end{tabular}%
    }
    \captionof{table}{\textbf{Scaling laws prediction errors.} We report the mean square error, R2 and mean absolute error for the loss prediction for held-in and held-out (8B model) data.}
    \label{tab:scaling_laws_errors_main}
\end{table}

\begin{table}[htb]
    \centering
    \setlength{\tabcolsep}{6pt} %
    \renewcommand{\arraystretch}{1.3} %
    \resizebox{0.9\linewidth}{!}{
    \begin{tabular}{lcccccccc}
        Model & E & $\alpha$ & $\beta$ & a & b  & d\\
        \midrule
        Avg  & 1.80922 & 0.29842 & 0.33209 & 0.54302  & 0.48301 &  0.92375 \\
        Std & 0.33811 & 0.10101 & 0.02892 & 0.08813 & 0.05787 & 0.23296 \\
        \bottomrule
    \end{tabular}%
    }
    \caption{\textbf{Scaling laws sensitivity.} We report the mean and standard deviation after bootstrapping with 100 iterations.}
    \label{tab:scaling_laws_sensitivity}
\end{table}

\subsection{Scaling laws evaluation}
\label{sec:scaling_laws_evaluation}
For each model size and number of training tokens, we compute the loss using the
estimated functional form in \cref{eq:scaling_laws} and compare it to the actual
loss observed in our runs. \Cref{fig:observed_vs_predicted_loss}, \Cref{fig:observed_vs_predicted_loss_extrapolation},
and \Cref{tab:scaling_laws_errors_main} visualizes these comparisons, showing
that our estimation is highly accurate, particularly for lower loss values and
larger FLOPs. We also assess our scaling laws in an extrapolation setting,
predicting performance beyond the model sizes used for fitting. Notably, our
approach estimates the performance of an 8B model with reasonable accuracy.  

Additionally, we conduct a sensitivity analysis using bootstrapping.
Specifically, we sample \( P \) points with replacement (\( P \) being the total
number of trained models) and re-estimate the scaling law coefficients. This
process is repeated 100 times, and we report the mean and standard deviation of
each coefficient. \Cref{tab:scaling_laws_sensitivity} shows that our
estimation is more precise for \(\beta\) than for \(\alpha\), primarily due to
the smaller number of model sizes relative to the number of different token
counts used to derive the scaling laws.

\begin{figure}[h!]
    \centering
    \captionsetup{type=figure}
    
    \begin{minipage}[t]{0.99\linewidth}
        \centering
        \input{graphs/early_late/pred_loss_vs_loss_early_extrapolation}
        \caption{\textbf{Observed vs predicted loss.} We visualize the loss predicted by our scaling laws \cref{eq:scaling_laws} and the actual loss achieved by each run. We can reliably predict the performance of models larger (8B params) than those used to fit the scaling laws.}
        \label{fig:observed_vs_predicted_loss_extrapolation}
    \end{minipage}
\end{figure}

\subsection{Scaling laws for sparse NMMs.}
\label{app:scaling_laws_moes}

Similar to dense models, we fit a parametric loss function (\Cref{eq:scaling_laws}) to predict the loss of sparse NMMs based on the number of parameters and training tokens, replacing the total parameter count with the number of active parameters. While incorporating sparsity is standard when deriving scaling laws for MoEs \citep{wangscalingmoe,krajewski2024scalingmoe,abnar2025parameters}, we focus on deriving scaling laws specific to the sparsity level used in our MoE setup. This yields coefficients that are implicitly conditioned on the sparsity configuration. 

We also experiment with a sparsity-aware formulation of the scaling law as proposed in \citep{abnar2025parameters}, and observe consistent trends (\Cref{tab:moes_coeffs}). In particular, the exponents associated with model size ($N$) are substantially larger than those for training tokens ($\beta$), reinforcing the importance of scaling model size in sparse architectures. Additionally, we observe that the terms governing the scaling of active parameters decompose into two components.

\input{tables/scaling_laws_coeffs_moes}

\section{Discussion and Limitations} 
\label{app:limitation}

\cpar{Scaling laws for multimodal data mixtures.} Our scaling laws study
spans different model configurations and training mixtures. While results
suggest that the scaling law coefficients remain largely consistent across
mixtures, a broader exploration of mixture variations is needed to validate this
observation and establish a unified scaling law that accounts for this factor.  

\cpar{Scaling laws and performance on downstream tasks.} Similar to
previous scaling law studies, our analysis focuses on pretraining performance as
measured by the validation loss. However, the extent to which these findings
translate to downstream performance remains an open question and requires
further investigation.

\cpar{Extrapolation to larger scales.} The accuracy of scaling law
predictions improves with increasing FLOPs~\cref{app:scaling_laws}.
Furthermore, we validate our laws when extrapolating to larger model sizes
(\cref{sec:scaling_laws_evaluation}). However, whether these laws can be reliably
extrapolated to extremely large model sizes remains an open question. 

\cpar{High resolution and early-fusion models.} Training early-fusion
models with high-resolution inputs leads to a significant increase in vision
tokens. While pooling techniques have been widely adopted for late-fusion
models, alternative approaches may be necessary for early fusion. Given
the similarity of early-fusion models to LLMs, it appears that techniques for
extending context length could be beneficial.

\cpar{Scaling laws for multimodal MoEs models.} For MoEs, we consider
only a single configuration (top-1 routing with 8 experts). We found this
configuration to work reasonably well in our setup, and follow a standard MoEs
implementation. However, the findings may vary when optimizing more the
MoE architecture or exploring different load-balancing, routing strategies
or different experts implementations.

\section{Mixture of experts and modality-specific specialization}
\label{app:moes}

\subsection{MoEs configuration}

We experiment with different MoEs configuration by changing the number of experts and the top-k. We report a sample of these experiments in \Cref{tab:moe_config}.

\begin{table*}[h!]
    \centering
    \setlength{\tabcolsep}{12pt} %
    \renewcommand{\arraystretch}{1} %
    \resizebox{0.9\linewidth}{!}{
    \begin{tabular}{lcccccccc}
        & \multicolumn{6}{c}{Accuracy} & \multicolumn{2}{c}{CIDEr}  \\
        \cmidrule(lr){2-7} \cmidrule(lr){8-9}
        & AVG & VQAv2 & TextVQA & OKVQA & GQA & VizWiz & COCO & TextCaps \\
        \shline
        4-E-top-1             & 40.0552&	64.068	&14.284 & 41.948& 61.46	&18.516& 62.201	& 34.08  \\
        8-E-top-1             & 41.6934&	65.684	&17.55	& 42.908& 63.26	&19.065& 67.877	& 39.63  \\
        8-E-top-2             & 42.8546&	66.466	&19.162 & 45.344& 63.94	&19.361& 65.988	& 41.649 \\
        8-E-top-2 finegrained & 39.904 &  62.76	&15.58	& 41.88	& 61.6  &17.7  & 57.52	& 35.42 \\
    \end{tabular}%
    }
    \caption{\textbf{SFT results with different MoEs configurations.} .}
    \label{tab:moe_config}
\end{table*}

\subsection{MoEs specialization}

\begin{figure}[h!]
    \centering
    \captionsetup{type=figure}
    \begin{subfigure}[t]{0.7\linewidth}
        \includegraphics[width=1.0\textwidth]{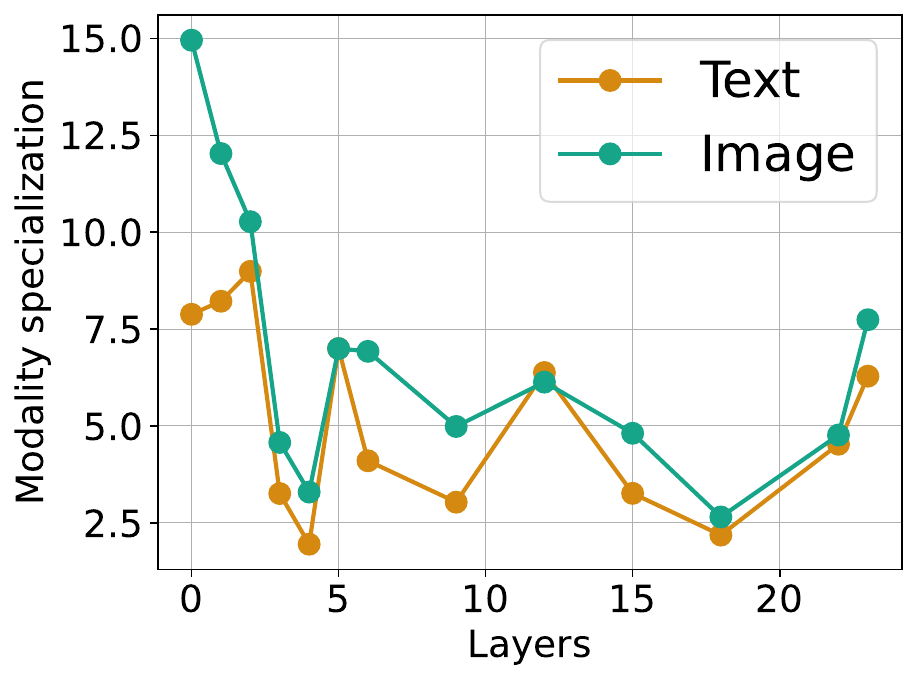}
    \end{subfigure}
    
    \caption{\textbf{Modality-specific specialization.} We visualize the experts specialization to text and image modalities. Models are evaluated on Obelics.}
    \label{fig:app_moes_specialization}
\end{figure}

We investigate multimodal specialization in MoE architectures. We compute a
specialization score as the average difference between the number of text/images
tokens assigned to each expert and a uniform assignment ($1/E$). Additionally,
we visualize the normalized number of text and image tokens assigned to each
expert across layers. \Cref{fig:app_moes_specialization} shows clear modality-specific
experts, particularly in the early layers. Furthermore, the specialization score
decreases as the number of layers increases but rises again in the very last
layers. This suggests that early and final layers require more modality
specialization compared to mid-layers. Additionally, we observe several experts
shared between text and image modalities, a phenomenon not present in
hard-routed or predefined modality-specific experts.

\begin{figure*}[h!]
    \centering
    \captionsetup{type=figure}
    \begin{subfigure}[t]{0.33\linewidth}
        \input{graphs/late/late_scaleflops_avg}
    \end{subfigure}
    \begin{subfigure}[t]{0.33\linewidth}
        \input{graphs/early/early_scaleflops_avg}
    \end{subfigure}
    \begin{subfigure}[t]{0.33\linewidth}
        \input{graphs/moe/moe_scaleflops_avg_big}
    \end{subfigure}

    \makebox[0.9\linewidth]{ %
        \begin{tikzpicture}
            \begin{axis}[
                hide axis, %
                xmin=0, xmax=0.5, ymin=0, ymax=1, %
                legend columns=6, %
                legend style={
                    at={(0.5, 1)}, %
                    anchor=north, %
                    /tikz/every even column/.append style={column sep=0.2cm}, %
                    scale=0.5, %
                    cells={align=left}, font=\footnotesize,
                },
            ]

            \addlegendimage{legend late_0_2b style}
            \addlegendentry{0.289B}
            \addlegendimage{legend late_0_4b style}
            \addlegendentry{0.494B}
            \addlegendimage{legend late_0_9b style}
            \addlegendentry{1B}
            \addlegendimage{legend late style}
            \addlegendentry{1.748B}
            \addlegendimage{legend late_2_2b style}
            \addlegendentry{2.430B}
            \addlegendimage{legend late_3_3b style}
            \addlegendentry{3.714B}

            \addlegendimage{legend early_0_2b style}
            \addlegendentry{0.275B}
            \addlegendimage{legend early_0_4b style}
            \addlegendentry{0.464B}
            \addlegendimage{legend early_0_9b style}
            \addlegendentry{0.932B}
            \addlegendimage{legend early style}
            \addlegendentry{1.627B}
            \addlegendimage{legend early_2_2b style}
            \addlegendentry{2.280B}
            \addlegendimage{legend early_3_3b style}
            \addlegendentry{3.354B}

            \addlegendimage{legend moe_0_2b style}
            \addlegendentry{0.275B}
            \addlegendimage{legend moe_0_4b style}
            \addlegendentry{0.464B}
            \addlegendimage{legend moe_0_9b style}
            \addlegendentry{0.932B}
            \addlegendimage{legend moe style}
            \addlegendentry{1.627B}
            \addlegendimage{legend moe_2_2b style}
            \addlegendentry{2.280B}
            \addlegendimage{legend moe_3_3b style}
            \addlegendentry{3.354B}

            \end{axis}
        \end{tikzpicture}
    }

    \vspace{-4cm}

    \caption{\textbf{Scaling laws for native multimodal models.} From left to right: late-fusion (dense), early-fusion (dense) and early-fusion MoEs. The scaling exponents are very close for all models. However, MoEs leads to overall lower loss (smaller multiplicative constant) and takes longer to saturate.}
    \label{fig:scaling_laws_early_late_moe}
\end{figure*}

\begin{figure*}[h!]
    \centering
    \captionsetup{type=figure}

    \begin{subfigure}[t]{0.33\linewidth}
        \input{graphs/late/late_scaleflops_getty}
    \end{subfigure}
    \begin{subfigure}[t]{0.33\linewidth}
        \input{graphs/late/late_scaleflops_obelics}
    \end{subfigure}
    \begin{subfigure}[t]{0.33\linewidth}
        \input{graphs/late/late_scaleflops_dclm}
    \end{subfigure}

    \begin{subfigure}[t]{0.33\linewidth}
        \input{graphs/early/early_scaleflops_getty}
    \end{subfigure}
    \begin{subfigure}[t]{0.33\linewidth}
        \input{graphs/early/early_scaleflops_obelics}
    \end{subfigure}
    \begin{subfigure}[t]{0.33\linewidth}
        \input{graphs/early/early_scaleflops_dclm}
    \end{subfigure}

    \begin{subfigure}[t]{0.33\linewidth}
        \input{graphs/moe/moe_scaleflops_getty}
    \end{subfigure}
    \begin{subfigure}[t]{0.33\linewidth}
        \input{graphs/moe/moe_scaleflops_obelics}
    \end{subfigure}
    \begin{subfigure}[t]{0.33\linewidth}
        \input{graphs/moe/moe_scaleflops_dclm}
    \end{subfigure}

    \makebox[0.9\linewidth]{ %
        \begin{tikzpicture}
            \begin{axis}[
                hide axis, %
                xmin=0, xmax=0.5, ymin=0, ymax=1, %
                legend columns=6, %
                legend style={
                    at={(0.5, 1)}, %
                    anchor=north, %
                    /tikz/every even column/.append style={column sep=0.2cm}, %
                    scale=0.5, %
                    cells={align=left}, font=\footnotesize,
                },
            ]

            \addlegendimage{legend late_0_2b style}
            \addlegendentry{0.289B}
            \addlegendimage{legend late_0_4b style}
            \addlegendentry{0.494B}
            \addlegendimage{legend late_0_9b style}
            \addlegendentry{1B}
            \addlegendimage{legend late style}
            \addlegendentry{1.748B}
            \addlegendimage{legend late_2_2b style}
            \addlegendentry{2.430B}
            \addlegendimage{legend late_3_3b style}
            \addlegendentry{3.714B}

            \addlegendimage{legend early_0_2b style}
            \addlegendentry{0.275B}
            \addlegendimage{legend early_0_4b style}
            \addlegendentry{0.464B}
            \addlegendimage{legend early_0_9b style}
            \addlegendentry{0.932B}
            \addlegendimage{legend early style}
            \addlegendentry{1.627B}
            \addlegendimage{legend early_2_2b style}
            \addlegendentry{2.280B}
            \addlegendimage{legend early_3_3b style}
            \addlegendentry{3.354B}

            \addlegendimage{legend moe_0_2b style}
            \addlegendentry{0.275B}
            \addlegendimage{legend moe_0_4b style}
            \addlegendentry{0.464B}
            \addlegendimage{legend moe_0_9b style}
            \addlegendentry{0.932B}
            \addlegendimage{legend moe style}
            \addlegendentry{1.627B}
            \addlegendimage{legend moe_2_2b style}
            \addlegendentry{2.280B}
            \addlegendimage{legend moe_3_3b style}
            \addlegendentry{3.354B}

            \end{axis}
        \end{tikzpicture}
    }

    \vspace{-4cm}
    \caption{\textbf{Scaling laws for native multimodal models.} From top to bottom: late-fusion (dense), early-fusion (dense) and early-fusion MoEs. From left to right: cross-entropy on the validation set of image-caption, interleaved and text-only data.}
    \label{fig:scaling_laws_early_late_moe_getty_obelics_dclm}
\end{figure*}

\begin{figure*}[h!]
    \centering
    \captionsetup{type=figure}
    \begin{subfigure}[t]{0.33\linewidth}
        \input{graphs/early/early_scaleflops_getty}
    \end{subfigure}
    \begin{subfigure}[t]{0.33\linewidth}
        \input{graphs/early/early_scaleflops_obelics}
    \end{subfigure}
    \begin{subfigure}[t]{0.33\linewidth}
        \input{graphs/early/early_scaleflops_dclm}
    \end{subfigure}

    \begin{subfigure}[t]{0.33\linewidth}
        \input{graphs/early_data_mixtures/early_scaleflops_getty_40_20_40}
    \end{subfigure}
    \begin{subfigure}[t]{0.33\linewidth}
        \input{graphs/early_data_mixtures/early_scaleflops_obelics_40_20_40}
    \end{subfigure}
    \begin{subfigure}[t]{0.33\linewidth}
        \input{graphs/early_data_mixtures/early_scaleflops_dclm_40_20_40}
    \end{subfigure}
    
    \begin{subfigure}[t]{0.33\linewidth}
        \input{graphs/early_data_mixtures/early_scaleflops_getty_30_30_40}
    \end{subfigure}
    \begin{subfigure}[t]{0.33\linewidth}
        \input{graphs/early_data_mixtures/early_scaleflops_obelics_30_30_40}
    \end{subfigure}
    \begin{subfigure}[t]{0.33\linewidth}
        \input{graphs/early_data_mixtures/early_scaleflops_dclm_30_30_40}
    \end{subfigure}

    \begin{subfigure}[t]{0.33\linewidth}
        \input{graphs/early_data_mixtures/early_scaleflops_getty_20_40_40}
    \end{subfigure}
    \begin{subfigure}[t]{0.33\linewidth}
        \input{graphs/early_data_mixtures/early_scaleflops_obelics_20_40_40}
    \end{subfigure}
    \begin{subfigure}[t]{0.33\linewidth}
        \input{graphs/early_data_mixtures/early_scaleflops_dclm_20_40_40}
    \end{subfigure}

    \makebox[0.9\linewidth]{ %
    \begin{tikzpicture}
        \node[anchor=north] (legend) at (0\linewidth, 0) {
            \begin{axis}[
                        hide axis, %
                        xmin=0, xmax=0.5, ymin=0, ymax=1, %
                        legend columns=6, %
                        legend style={
                            at={(-0.12, -0.025)}, %
                            anchor=north, %
                            /tikz/every even column/.append style={column sep=0.2cm}, %
                            scale=0.5,
                            cells={align=left}, font=\footnotesize,
                            anchor=center,
                        },
                    ]
                \addlegendimage{legend early_0_2b style}
                \addlegendentry{0.275B}
                \addlegendimage{legend early_0_4b style}
                \addlegendentry{0.464B}
                \addlegendimage{legend early_0_9b style}
                \addlegendentry{0.932B}
                \addlegendimage{legend early style}
                \addlegendentry{1.627B}
                \addlegendimage{legend early_2_2b style}
                \addlegendentry{2.280B}
                \addlegendimage{legend early_3_3b style}
                \addlegendentry{3.354B}
            \end{axis}
        };
    \end{tikzpicture}
    }
    \vspace{0.5cm}

    \caption{\textbf{Scaling laws for early-fusion native multimodal models.} Our runs across different training mixtures (Image-caption-Interleaved-Text) and FLOPs. We visulize the final validation loss on 3 data types: HQITP (left), Obelics (middle) and DCLM (right).}
    \label{fig:app_early_scaleflops_data_mixtures}
\end{figure*}

%% file: figs/late_vs_early_equal_tokens.tex
\begin{figure*}[t!]
    \centering
    \captionsetup{type=figure}
    \begin{subfigure}[t]{0.33\linewidth}
        \input{graphs/early_late/early_vs_late_scaledata_getty}
    \end{subfigure}
    \begin{subfigure}[t]{0.33\linewidth}
        \input{graphs/early_late/early_vs_late_scaledata_obelics}
    \end{subfigure}
    \begin{subfigure}[t]{0.33\linewidth}
        \input{graphs/early_late/early_vs_late_scaledata_dclm}
    \end{subfigure}
    \vspace{-15pt}
        \begin{center}
            \ref{sharedlegend}
        \end{center}
    \caption{\textbf{Early vs late fusion: scaling training FLOPs.} We compare
    early and late fusion models when scaling both the model size and the number
    of training tokens. The gap decreases mainly due to scaling models size.}
    \label{fig:early_vs_late_scaledata_main}
\end{figure*}

%% file: figs/early_vs_late_imageres.tex
\begin{figure}[t!]
    \centering
    \captionsetup{type=figure}
    \begin{subfigure}[t]{0.49\linewidth}
        \input{graphs/early_late/early_vs_late_imageres_getty}
    \end{subfigure}
    \begin{subfigure}[t]{0.49\linewidth}
        \input{graphs/early_late/early_vs_late_imageres_obelics}
    \end{subfigure}
    \vspace{-7pt}
    \caption{\textbf{Early vs late fusion: training with different image
    resolutions (isoFLOPs).} For the same training FLOPs we vary the image
    resolution (and thus the number of image tokens) during training and report
    the final training loss. Increasing resolution, hurts the performance on
    text and interleaved documents, while helping image captioning. The gap
    stays almost the same on text and interleaved data while slightly increase
    on image captioning in favor of early fusion.}
    \label{fig:early_vs_late_imageres}
\end{figure}

%% file: figs/early_vs_late_datatype_isoparams.tex
\begin{figure*}[t!]
    \centering
    \captionsetup{type=figure}
    \begin{subfigure}[t]{0.32\linewidth}
         \input{graphs/early_late/early_late_datatype_getty}
    \end{subfigure}
    \begin{subfigure}[t]{0.32\linewidth}
        \input{graphs/early_late/early_late_datatype_obelics}
    \end{subfigure}
    \begin{subfigure}[t]{0.32\linewidth}
         \input{graphs/early_late/early_vs_late_datatype_dclm}
    \end{subfigure}
    
    \makebox[0.9\linewidth]{ %
        \begin{tikzpicture}
            \begin{axis}[
                hide axis, %
                xmin=0, xmax=0.5, ymin=0, ymax=1, %
                legend columns=4, %
                legend style={
                    at={(0.5, 1)}, %
                    anchor=north, %
                    /tikz/every even column/.append style={column sep=0.2cm}, %
                    scale=0.5, %
                    cells={align=left}, font=\footnotesize,
                },
            ]

                \addlegendimage{legend late style}
                \addlegendentry{L}

               \addlegendimage{EarlyGradStart!10!EarlyGradEnd, thick, solid, mark=*, mark size=1.75pt}
                \addlegendentry{E (Text)}

                \addlegendimage{legend early style, mark size=1.75pt}
                \addlegendentry{E (FLOPs)}

               \addlegendimage{EarlyGradStart!95!EarlyGradEnd, thick, solid, mark=*, mark size=1.75pt}
                \addlegendentry{E (Params)}
            \end{axis}
        \end{tikzpicture}
    }
    
    \vspace{-5cm}
    
    \caption{\textbf{Early vs late fusion: changing the training mixture and early-fusion configuration.} We
    vary the training mixtures and plot the final training loss for different
    configuration of early fusion models. For the
    same number of total parameters early fusion consistently outperform late
    fusion.}
    \label{fig:early_vs_late_datatype_isoparams}
\end{figure*}

%% file: figs/early_vs_late_init_scaledata.tex
\begin{figure}[t!]
    \centering
    \captionsetup{type=figure}
    \begin{subfigure}[t]{0.32\linewidth}
        \input{graphs/early_late_init/early_vs_late_init_scaledata_getty}
    \end{subfigure}
    \begin{subfigure}[t]{0.32\linewidth}
        \input{graphs/early_late_init/early_vs_late_init_scaledata_obelics}
    \end{subfigure}
    \begin{subfigure}[t]{0.32\linewidth}
        \input{graphs/early_late_init/early_vs_late_init_scaledata_dclm}
    \end{subfigure}

    \makebox[0.9\linewidth]{ %
        \begin{tikzpicture}
            \begin{axis}[
                hide axis, %
                xmin=0, xmax=0.5, ymin=0, ymax=1, %
                legend columns=2, %
                legend style={
                    at={(0.5, 1)}, %
                    anchor=north, %
                    /tikz/every even column/.append style={column sep=0.2cm}, %
                    scale=0.5, %
                    cells={align=left}, font=\footnotesize,
                },
            ]
            \addlegendimage{LateGradStart!75!LateGradEnd, thick, solid, mark=*, mark size=1.5pt}
            \addlegendentry{Late-init}

            \addlegendimage{legend early_2_2b style}
            \addlegendentry{Early-Init}
            \end{axis}
        \end{tikzpicture}
    }
    
    \vspace{-5cm}
    \caption{\textbf{Early vs late fusion when initializing the encoder and decoder.} Early-fusion can match the performance of late-fusion models when trained for longer. However, the gap is bigger on image-caption data.}
    \label{fig:early_vs_late_init_scaledata}
\end{figure}

%% file: tables/scaling_laws_coeffs_moes.tex
\begin{table*}[h!]
    \centering
    \begin{minipage}[b]{0.9\linewidth}
        \centering
        \setlength{\tabcolsep}{24pt}
        \renewcommand{\arraystretch}{1}
        \resizebox{1\linewidth}{!}{
        \begin{tabular}{ccc}
             \grayrow $L(N, D) = E+\frac{A}{N^{\alpha}}+\frac{B}{D^{\beta}}$ &vs&  $L(N, D, S) = \frac{A}{N^\alpha} + \frac{B}{D^\beta} + \frac{C}{(1 - S)^\lambda} + \frac{d}{(1 - S)^\delta N^\gamma} + E$ \\
        \end{tabular}
        }
      \label{tab:power_laws}
    \end{minipage}
    \begin{minipage}[b]{0.9\linewidth}
        \centering
        \setlength{\tabcolsep}{8pt}
        \renewcommand{\arraystretch}{1.1}
        \resizebox{1\linewidth}{!}{
        \begin{tabular}{lcccccccccc}
            Model & E & A & B & $\alpha$ & $\beta$ & $\lambda$ & $\delta$  & $\gamma$ & C & d  \\ %
            \shline
            
            $L(N, D)$ (\Cref{eq:scaling_laws}) & 2.158  & 381773 & 4659 & 0.710 & 0.372 & --  & -- & -- & -- & -- \\
            $L(N, D, S)$ \citep{abnar2025parameters} &  1.0788 & 1 & 4660 & 0.5890 & 0.3720 & 0.2 & 0.2 & 0.70956 & 1.0788 & 381475
        \end{tabular}%
        } 
        \caption{\textbf{Scaling laws for sparse native multimodal models}.}
        \label{tab:moes_coeffs}
    \end{minipage}    
\end{table*}